\documentclass{ieeetmlcn}
\usepackage{algorithmic}
\usepackage{graphicx,color}
\usepackage{textcomp}
\usepackage{xcolor}
\usepackage{hyperref}
\hypersetup{hidelinks=true}
\def\BibTeX{{\rm B\kern-.05em{\sc i\kern-.025em b}\kern-.08em
    T\kern-.1667em\lower.7ex\hbox{E}\kern-.125emX}}
\AtBeginDocument{\definecolor{tmlcncolor}{cmyk}{0.93,0.59,0.15,0.02}\definecolor{NavyBlue}{RGB}{0,86,125}}

\makeatletter
\newcommand{\seclogo}{}
\makeatother

\def\authorrefmark#1{\ensuremath{^{\textbf{#1}}}}

\usepackage{graphicx}
\usepackage{epsf}
\usepackage{epsfig}
\usepackage{epstopdf}
\usepackage{ifpdf}
\usepackage{cite}
\usepackage[utf8]{luainputenc}
\usepackage{amssymb}
\usepackage{amsthm}
\usepackage{lipsum}
\usepackage{bbm}

\usepackage{pifont}

\usepackage[cmex10]{amsmath}
\usepackage{amsmath}
\usepackage{amssymb}
\usepackage{dsfont} 
\DeclareMathOperator*{\argmax}{arg\,max} 
\usepackage{algorithmic}
\usepackage{array}
\usepackage{mdwmath}
\usepackage{mdwtab}
\usepackage{eqparbox}
\usepackage{xcolor}
\usepackage{fixltx2e}
\usepackage{stfloats}
\usepackage{url}

\usepackage{breakurl}
\usepackage{subcaption}
\usepackage{amsfonts}
\usepackage{bbm}
\usepackage{breakurl}
\usepackage{doi}
\usepackage{float}
\usepackage{epsf}
\usepackage{epsfig}
\usepackage{epstopdf}
\usepackage{algorithmic}
\usepackage[linesnumbered, ruled, vlined]{algorithm2e}
\SetKw{KwOr}{or}
\SetKw{KwAnd}{and}
\SetKw{Cnt}{continue}

\IEEEoverridecommandlockouts

\begin{document}
\receiveddate{XX Month, XXXX}
\reviseddate{XX Month, XXXX}
\accepteddate{XX Month, XXXX}
\publisheddate{XX Month, XXXX}
\currentdate{XX Month, XXXX}
\doiinfo{TMLCN.2022.1234567}

\markboth{}{Lotfi {et al.}}

\title{Task-Specific Sharpness-Aware O-RAN Resource Management using Multi-Agent Reinforcement Learning}

\author{Fatemeh Lotfi\authorrefmark{1}, Hossein Rajoli\authorrefmark{1},\\ and Fatemeh Afghah\authorrefmark{1}, Senior Member, IEEE}
\affil{Holcombe Department of Electrical and Computer Engineering, Clemson University, Clemson, SC, 29631, USA }
\corresp{Corresponding author: Fatemeh Lotfi (email: flotfi@clemson.edu).}
\authornote{This material is based upon work supported by the National Science Foundation under Grant Numbers  CNS-2202972, CNS- 2318726, and CNS-2232048.}




\begin{abstract}
Next-generation networks utilize the Open Radio Access Network (O-RAN) architecture to enable dynamic resource management, facilitated by the RAN Intelligent Controller (RIC). While deep reinforcement learning (DRL) models show promise in optimizing network resources, they often struggle with robustness and generalizability in dynamic environments. 
This paper introduces a novel resource management approach that enhances the Soft Actor Critic (SAC) algorithm with Sharpness-Aware Minimization (SAM) in a distributed Multi-Agent RL (MARL) framework. 
Our method introduces an adaptive and selective SAM mechanism, where regularization is explicitly driven by temporal-difference (TD)-error variance, ensuring that only agents facing high environmental complexity are regularized. 
This targeted strategy reduces unnecessary overhead, improves training stability, and enhances generalization without sacrificing learning efficiency. 
We further incorporate a dynamic $\rho$ scheduling scheme to refine the exploration-exploitation trade-off across agents. Experimental results show our method significantly outperforms conventional DRL approaches, yielding up to a $22\%$ improvement in resource allocation efficiency and ensuring superior QoS satisfaction across diverse O-RAN slices.
\end{abstract}

\begin{IEEEkeywords}
Open RAN, Deep reinforcement learning, Network slicing and scheduling, Sharpness aware minimization, Multi agent reinforcement learning. \vspace{-0.cm}
\end{IEEEkeywords}


\maketitle

\section{Introduction}\vspace{-0.cm}

\IEEEPARstart{A}{dvanced} 5G/6G networks, especially those utilizing open radio access network (O-RAN) architecture, are designed to be highly dynamic and flexible. This allows operators to manage network slices on-the-fly as user demands and network conditions evolve. Such dynamic management of network slices is a key feature of 5G/6G and O-RAN, enabling operators to maintain adaptability and responsiveness to changing demands and diverse use cases~\cite{alam2024comprehensive,3gppRe18,polese2022understanding}. Furthermore, network resource management plays a crucial role in these modern networks, providing substantial benefits through dynamic resource allocation that allows for modifications to network slices without service disruptions.

To manage resources in wireless networks, the O-RAN architecture offers an opportunity to effectively employ artificial intelligence (AI) and machine learning (ML) approaches for network operations, particularly through its RAN Intelligent Controller (RIC) modules~\cite{alam2024comprehensive,polese2022understanding, 3gppRe18}. These modules not only enhance network functionalities by enabling intelligent resource management and sophisticated control techniques but are also pivotal in delivering advanced services due to their capacity to facilitate real-time data collection and analysis between multiple disaggregated Central Units (CUs) and Distributed Units (DUs)~\cite{alam2024comprehensive,polese2022understanding}. This dynamic and flexible framework supports the integration of deep reinforcement learning (DRL) approaches, which can efficiently train decision-making policies to adapt to the constantly changing wireless network environment. By leveraging DRL within the O-RAN architecture, it is possible to optimize network performance, improve resource allocation, and enhance the overall user experience through more effective and responsive network management~\cite{raftopoulos2024drl,rezazadeh2022specialization,polese2022understanding, thaliath2022predictive, cheng2022reinforcement, lotfi2022evolutionary, chen2023hierarchical, lotfiattention, rezazadeh2021collaborative,kim2019reinforcement,lotfi2024open,yang2022multi,zhang2022team,lotfi2025llm,lotfi2025meta,zhang2022federated,alsenwi2021,ebrahimi2025intelligent,lotfi2025oran,lotfi2025prompt}. 
Specifically, the optimization challenge we address involves the strategic allocation of network slices and Resource Blocks (RBs) within the O-RAN architecture, modeled as a distributed Multi-Agent Reinforcement Learning (MARL) problem. Each agent, representing a DU, is tasked with dynamically managing resources to address the dynamism of wireless networks and their requirement for rapid adaptation to unexpected conditions. This MARL setup forms an integral part of the broader DRL strategies employed to enhance network performance. The decentralized nature of this model is essential because it allows each DU agent to independently adapt to local network conditions and user demands, enhancing responsiveness and scalability. 

Among DRL methods, policy-based actor-critic methods, such as the Soft Actor Critic (SAC), stand out for their exceptional performance in dynamic cellular networks. SAC optimally balances exploration and exploitation through entropy maximization, enabling robust policy learning that can navigate the complexities of wireless networks. This capability allows network operators to achieve enhanced resource allocation and improved overall network performance, making SAC an ideal fit for the challenges of dynamic environments. 
Although DRL approaches are impressive in dynamic network slicing and resource allocation, they often suffer from model overfitting to earlier interactions and fail to adapt to new experiences~\cite{li2017deep,zhang2018study}. Model generalization is a crucial concept that the literature focuses on~\cite{zhang2023federated,petzka2021relative,zhao2022penalizing}. Traditional methods like L2-regularization~\cite{liu2020regularization}, which penalizes large weights to promote simpler models, have been used to mitigate overfitting but often require precise tuning and may not fully adapt to the rapidly changing environments typical of wireless networks. 


\begin{figure*}[t!]
  \centerline{\includegraphics[width=4.5in]{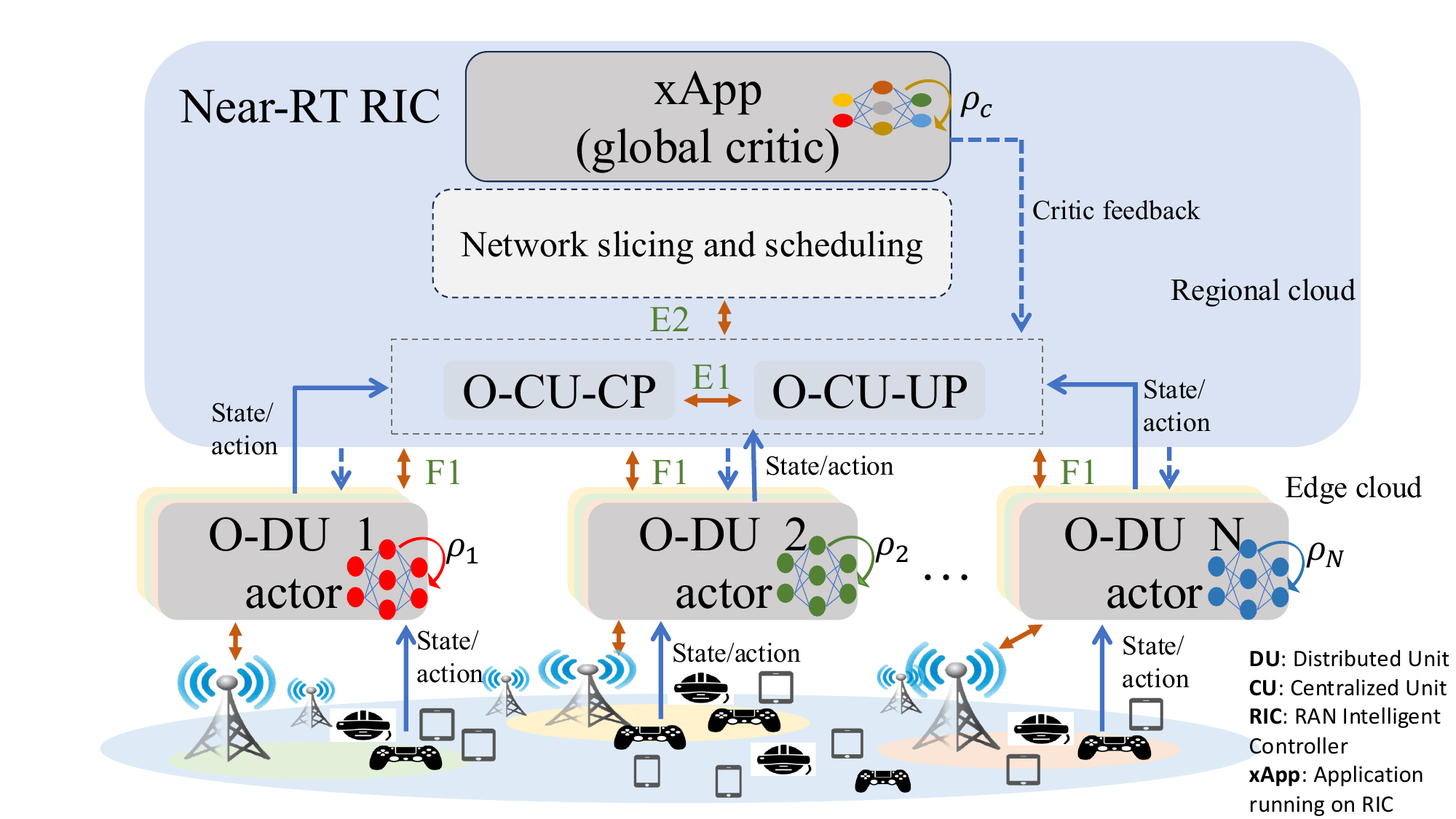}}
    \caption{\small System topology showing multi-agent Soft Actor-Critic (SAC) architecture for O-RAN. Each O-DU serves as a local actor, while the xApp in the near-RT RIC functions as the global critic. Solid arrows indicate action/state flow from actors to the critic; dashed feedback loops represent critic updates to the actors. The architecture interfaces with the O-CU via E1 and the near-RT RIC via E2. 
    }
    \label{sys_graph}
    \vspace{-0.cm}
\end{figure*}
To address these limitations more effectively, the Sharpness-Aware Minimization (SAM) optimizer has been proposed in literature~\cite{foret2020sharpness,lee2024plastic}. Unlike conventional approaches, SAM optimizes the worst-case loss near the current parameters, smoothing the loss landscape to promote stable and generalizable learning. 
This stability makes SAM particularly suitable for the dynamic nature of wireless networks, where it enhances generalization and robustness to ensure consistent performance in dynamic network slicing and resource allocation scenarios that require continuous adaptation~\cite{lee2024plastic,yi2022improved}. 
Another important consideration in DRL approaches is balancing exploration and exploitation. In this paper, we propose a novel approach as Task-Aware SAM (TA-SAM) within MARL framework to effectively manage this balance. A key novelty of our approach is the use of temporal-difference (TD)-error variance as a proxy for agent-level learning uncertainty, allowing SAM to be selectively applied only to agents encountering unstable learning conditions. 
The dynamic TA-SAM MARL approach allows for the control of exploration-exploitation balance through a hyperparameter $\rho$. Unlike standard regularization techniques, SAM directly shapes the loss landscape by smoothing it, thus promoting stable learning and enhancing model generalization. 

While prior work such as \cite{lee2024plastic} demonstrates the potential of techniques like SAM in enhancing generalization and sample efficiency in single-agent RL settings, its scope is limited to controlled environments. It does not address the challenges of multi-agent systems or dynamic real-world applications. Our work bridges this gap by integrating SAM into a distributed MARL framework tailored for O-RAN environments. By dynamically adjusting SAM’s hyper-parameters based on environmental complexity, we enable robust learning and effective resource allocation in highly dynamic network scenarios. 
\noindent The main contributions of this paper are summarized as follows: 
\begin{itemize}
    \item We propose TA-SAM, a novel task-aware framework that incorporates SAM within a distributed MARL setup for joint network slicing and resource allocation in O-RAN. Unlike single-agent baselines (e.g., PLASTIC), our design supports distributed training across multiple DUs and UE-level granularity.
    
    \item
    We are the first to introduce a TD-error driven selective application of SAM to actor networks. By leveraging the TD-error as a proxy for task complexity, our method dynamically regulates the sharpness penalty, improving policy robustness and learning efficiency under non-stationary conditions.
    
    \item We develop a dynamic $\rho$ adjustment mechanism that adapts the SAM neighborhood radius based on the evolving environment. This mechanism refines the exploration-exploitation trade-off and enables adaptive learning in heterogeneous wireless settings.
    
    \item We show that incorporating SAM into the global critic model within a distributed MARL framework leads to better generalization across distributed actors, enhancing collaborative decision-making and stability across varied traffic and QoS profiles.
    
    \item We tailor the SAC algorithm to support discrete RB allocation actions in a continuous state space by applying a sigmoid activation to the actor’s output, followed by a thresholding step to convert continuous values into binary allocation decisions, thereby preserving SAC’s optimization benefits in this hybrid setting.
    
    \item Through extensive experiments, we demonstrate that TA-SAM improves resource allocation efficiency and user QoS by up to 22\%, outperforming conventional DRL and task-adaptive baselines under diverse traffic and network load scenarios.
\end{itemize}
\noindent \emph{To the best of our knowledge, this is the first work that adaptively integrates the SAM optimizer into MARL setting for O-RAN jointly slicing and scheduling}.

The structure of this paper is outlined as follows: Section \ref{sec:relatedworks} reviews related literature in the field. Section\ref{sec:sysmodel} details the system model and formulates the problem for joint slicing and scheduling in O-RAN. Section \ref{sec:SAM-MARL} introduces our proposed TA-SAM MARL approach for jointly tackling slicing and scheduling, including a description of the dynamic TA-SAM MARL algorithm designed to address the MDP problem. Simulation results are discussed in Section \ref{sec:simulation}, followed by conclusions in Section \ref{sec:conclusion}.\vspace{-0.cm}

\section{Related Works}\label{sec:relatedworks}\vspace{-0.cm}
\subsection{DRL-based network slicing and scheduling}\vspace{-0.cm}
The integration of AI and ML techniques, including DRL, offers significant advantages for network slicing within the O-RAN architecture. However, these approaches encounter several challenges, such as the requirement for extensive and diverse datasets and the necessity for thorough exploration to accelerate convergence. These challenges need to be addressed to develop ML models that are robust across various scenarios without negatively impacting RAN performance. Recent literature has focused on these challenges, exploring the complexities of O-RAN network slicing and proposing innovative DRL-based solutions to overcome the inherent difficulties~\cite{polese2022understanding, thaliath2022predictive, cheng2022reinforcement, lotfi2022evolutionary, chen2023hierarchical, lotfiattention,rezazadeh2021collaborative,kim2019reinforcement, lotfi2024open,lotfi2025llm,lotfi2025oran,lotfi2025prompt,lotfi2025meta}. 

Addressing these challenges, \cite{cheng2022reinforcement} introduces a RL-based method that intelligently allocates resources across different network slices using Q-learning. By utilizing a shared Q-table across DU and CU, the algorithm optimizes resource allocation within a system that includes a single CU and multiple DUs. This method not only improves resource distribution but also enhances the coordination between various network modules. Exploring alternative approaches, \cite{lotfi2022evolutionary} combines evolutionary algorithms with DRL to promote adaptability in the learning process itself, allowing the system to evolve and better handle unpredictable network conditions. Meanwhile, \cite{chen2023hierarchical} employs a hierarchical RL framework that, while benefiting from structured efficiency, may encounter challenges with irregular network demands, which could increase computational requirements. The integration of attention mechanisms within DRL, as explored in \cite{lotfiattention}, represents an advancement in context-aware resource allocation, effectively improving the responsiveness of network slicing to real-time conditions. Complementing this, \cite{rezazadeh2021collaborative} introduces a Collaborative Statistical Actor-Critic (CS-AC) framework within a 6G-like RAN setting, leveraging DRL to manage slice performance proactively and on a large scale. This approach, by employing multiple actors and a single global critic, improves both the exploration-exploitation balance and the scalability of network slicing operations. Building on this concept, \cite{kim2019reinforcement} investigates the application of actor-critic methods for dynamic resource management in network slicing, illustrating how DRL can foster enhanced balance and generalizability in managing network resources effectively. Similarly, \cite{lotfi2024open} extends predictive capabilities by integrating LSTM networks with DRL to forecast traffic patterns and proactively adjust resource distribution, a crucial strategy for maintaining service quality during peak periods.\vspace{-0.cm}
Additionally, DRL has been effectively applied in related communication problems under uncertainty, such as scheduling decisions in remote traffic estimation scenarios~\cite{peng2022communication}, demonstrating its versatility in managing complex resource-constrained environments. 
\subsection{MARL-based network optimization}
Collaborative strategies in multi-agent systems have proven crucial in enhancing the management of dynamic wireless networks, offering substantial benefits in adaptability and resource optimization~\cite{yang2022multi}. 
Building on collaboration approaches, authors in \cite{zhang2022team} introduce a cooperative strategy involving two distinct network controllers, designed to optimize system throughput. To facilitate effective collaboration and avoid potential conflicts, the actions taken by one controller are integrated as the state inputs for the other one, ensuring a harmonized operation. The authors in \cite{zhang2022federated} expand on this idea by implementing a federated RL system where multiple controllers share insights to improve decision-making, which significantly boosts the network's efficiency. This method employs Deep Q-learning for each individual controller. During the federated learning phase, Q-tables from each controller are amalgamated to formulate a comprehensive global model, which then guides action selection across the network. Furthermore, \cite{lotfi2022evolutionary,lotfiattention,lotfi2024open,lotfi2025meta,lotfi2025prompt,lotfi2025llm,lotfi2025oran} demonstrate the application of MARL strategies in network slicing models. These papers highlight how multi-agent systems adeptly handle the complexities of real-time network adjustments. Specifically, \cite{lotfi2022evolutionary} employs a unique approach where multiple agents, functioning like individual entities in a genetic-based evolutionary algorithm, leverage their collective experiences to enhance environmental awareness. Similarly, \cite{lotfiattention,lotfi2024open,lotfi2025meta,lotfi2025prompt,lotfi2025llm,lotfi2025oran} utilize the dispersed trajectories of agents' experiences across the wireless network to gain a comprehensive understanding of the environment, which significantly informs the network's dynamic updates. 

Furthermore, \cite{naderializadeh2021resource} explores the deployment of MARL in multi-cell wireless networks to address challenges related to user scheduling and downlink power control. This method equips each transmitter with a deep RL agent that makes decisions based on delayed observations and information exchanged with neighboring agents. This framework enhances distributed decision-making among agents who are unaware of each other's actions, thereby improving scalability and adaptability to different network sizes and configurations. Additionally, \cite{rezazadeh2022specialization} addresses the challenges of managing network resources in a scalable and distributed manner within 6G RAN through the use of Federated Deep Reinforcement Learning (FDRL). It introduces a hierarchical architecture that facilitates efficient resource orchestration across multiple network slices by employing traffic-aware local decision agents within the RAN. These agents dynamically adapt their resource allocation policies based on real-time traffic conditions and operate under a federated learning framework to enhance decision-making efficiency without relying on a centralized controller.\vspace{-0cm}

The PLASTIC framework \cite{lee2024plastic} leverages SAM to enhance input and label plasticity, achieving competitive performance on benchmarks such as Atari-100k and DeepMind Control Suite. However, its focus remains on single-agent systems in static environments, with no exploration of distributed or multi-agent scenarios. In contrast, in this work, we incorporate Task-Aware SAM (TA-SAM) into a MARL framework for O-RAN, addressing the unique challenges of decentralized decision-making, dynamic network slicing, and scheduling. 
Unlike conventional SAM, our approach selectively applies sharpness-aware optimization to specific actors, targeting those encountering the most complex and unstable learning conditions. Furthermore, we introduce a novel dynamic hyperparameter adjustment mechanism that optimizes the selective application of SAM in real time and enhances generalization and stability in dynamic network environments. 
In parallel, the challenge of generalization in DRL has received increasing attention. It is well-established that DRL agents often overfit to narrow training scenarios and fail to generalize across dynamic environments \cite{packer2019assessing}. Recent studies in loss landscape analysis also suggest that the geometry of the solution, particularly the flatness or sharpness of minima, correlates with generalization ability \cite{li2018sharp}. Our work integrates these perspectives by leveraging SAM to guide policy learning toward flatter optima, improving robustness in non-stationary O-RAN settings.


\section{SYSTEM MODEL AND PROBLEM FORMULATION}\label{sec:sysmodel}

\subsection{Overall Architecture}
In an O-RAN  network, the architecture is segmented into $|\mathcal{L}|=3$ distinct network slices, each designed to serve the specific service requirements. These slices include Enhanced Mobile Broadband (eMBB) for high-speed data services, Massive Machine-Type Communications (mMTC) for connecting a large number of devices, and Ultra-Reliable Low-Latency Communications (URLLC) for critical applications requiring immediate response. Each slice has unique Quality of Service (QoS) requirements. In this architecture, we consider several CUs, DUs, and RUs modules. These elements are coordinated by a RIC module to dynamically manage system resources. This dynamic nature involves two key resource management steps: slicing, to allocate network resources to different services, and scheduling, to distribute these resources among individual users within a slice. The inter-slice manager handles resource management across different network slices, while the intra-slice manager manages resources within a single slice. The network's goal is to satisfy these diverse QoS demands by intelligently controlling resources between slices through the RIC module. Subsequently, the medium access control (MAC) layer must assign resources to UE devices based on the resource allocation strategy outlined by the slice management. 
To tackle this, we introduce our wireless model and formulate the slice management problem, considering wireless resource constraints. Fig. \ref{sys_graph} illustrates our proposed scenario. \vspace{-0.cm}
\subsection{Achievable Data Rate}

The O-RAN architecture illustrated in Fig. \ref{sys_graph} employs various network slices, each with different QoS criteria, to serve diverse services for UEs. The RIC module oversees all slices and resources, utilizing dynamic resource management to adjust to network dynamics and fluctuating wireless channels. However, the stochastic nature of this dynamic approach complicates resource allocation. To mitigate intra-cell interference, an orthogonal frequency-division multiplexing (OFDM) transmission scheme is employed~\cite{3gpp15}. The achievable data rate for each UE $u$, associated with slice $l$ and assigned to RU-DU $m$
, considering a Rayleigh fading transmission channel and assuming the system operates within discrete time frames $t$, can be expressed as:\vspace{-0.cm}
\begin{align}\label{urate} 
    c^l_{u,m}(t) = \sum_{k=1}^{K^l_{m}} &B e_{u,k} b_{l,k} \log \Big(1+\nonumber\\
    &\frac{p_{k,m}(t) d_{u,m}(t)^{-\eta} |h_{u,k}(t)|^2}{ I_{u,k}(t)+ \sigma^2}\Big),
\end{align}
where $e_{u,k} \in \{0, 1\}$ is a binary variable indicating RB allocation for user $u$ in RB $k$,  $b_{l,k}$ is RB distribution indicator for slice $l$ in RB $k$, and $B$ represents reach RB bandwidth. 
Let $p_{k,m}$ denote the transmission power allocated to resource block $k$ of RU $m$ and let $K^l_{m}$ represent the total available resource blocks of RU $m$ that belongs to slice $l$. The variable $d_{u,m}(t)$ represents the distance between RU $m$ and UE $u$ at each time frame, which varies over time due to the mobility of the UEs. Additionally, $\eta$ represents the path loss exponent, and $|h_{u,k}(t)|^2$ indicates the time-varying channel gain due to Rayleigh fading for each time frame. In equation \eqref{urate}, the term $I_{u,k}(t)=\sum_{m' \neq m} \sum_{u'\neq u} e_{u',k} p_{k,m'}(t) d_{u',m'}(t)^{-\eta} |h_{u',k}(t)|^2$, indicates the inter-cell interference of downlink transmission from other RUs on RB $k$, and  $\sigma^2$ is the additive white Gaussian noise (AWGN) variance.

\subsection{QoS for Network Slices}
We consider a set of key performance indicators (KPIs) 
$\mathcal{M}=\{l_d, \mu_r, s_a, d_s\}$ covering latency, 
throughput, service availability, and user density support. 
Each slice applies priority weights $\omega_{i,l}$ to emphasize 
relevant metrics. 
The latency metric is computed as:
\begin{align}
l_d = \max(\tau_i), \quad \forall i \in N_u,
\end{align} 
The average throughput metric is:
\begin{align}
\mu_r = \frac{1}{N_u}\sum_{i=1}^{N_u} C_i,
\end{align} 
The service-availability metric is:
\begin{align}
s_a = \frac{1}{T}\sum_{t=1}^{T}\mathds{1}_{(C_i(t)\ge \lambda_i)}, \quad \forall i\in N_u,
\end{align} 
The user-density support metric is defined as:
\begin{align}
d_s = \frac{1}{N_u}\sum_{i=1}^{N_u}\mathds{1}_{(C_i \ge \lambda_i)}.
\end{align} 
Finally, slice-level QoS is computed as:
\begin{align}
Q^{l} = \sum_{i=1}^{|\mathcal{M}|} \omega_{i,l}\mathcal{M}_i.
\end{align}

\subsection{Problem Formulation}
The objective is to maximize the network's overall performance as $\Omega(\{Q^l |\forall l \in \mathcal{L}\})$ which is a function that appropriately aggregates distinct $Q^l$ as an overall network performance while ensuring the specific QoS requirements for each slice are met. The problem defined in \eqref{opt1} jointly optimizes the QoS for all slices in the network, ensuring that each slice’s service-level constraints are respected while maximizing the overall system efficiency. 
This is represented as: \vspace{-0.cm}
\begin{subequations} 
\begin{align}\label{opt1}
 \argmax_{\boldsymbol{b},\boldsymbol{e}} & \hspace{0.5cm} 
 \Omega\big(\{Q^l(\boldsymbol{b},\boldsymbol{e})|\forall l \in \mathcal{L}\}\big),\\
 \text{s.t.,} 
& \hspace{0.5cm}  \sum_{l=1}^{|\mathcal{L}|}\sum_{u=1}^{N_{u,m}}\sum_{k=1}^{K_m}e_{u,k}b_{l,k} \leq K_m, \label{opt1_rb}\\
& \hspace{0.5cm} \sum_{l}b_{l,k} \leq 1,\,\,   \label{opt1_e}\\
& \hspace{0.5cm} Q^l_{\boldsymbol{b},\boldsymbol{e}}\geq Q^l_{min}, \label{opt1_qos} \vspace{-0.cm}
\end{align}
\end{subequations}
\noindent
where:
\begin{itemize}
    \item $\Omega(\cdot)$: overall network performance aggregation function across slices,
    \item $Q^l(\boldsymbol{b}, \boldsymbol{e})$: QoS metric for slice $l$ under allocation $\boldsymbol{b}, \boldsymbol{e}$,
    \item $\boldsymbol{b} \in \{0,1\}^{|\mathcal{L}| \times K^l_m}$: binary RB assignment for each slice $l$ over RBs,
    \item $\boldsymbol{e} \in \{0,1\}^{N_{u,m} \times K^l_m}$: binary RB-user assignment matrix,
    \item $K_m$: total number of RBs available at RU-DU $m$,
    \item $Q^l_{min}$: minimum QoS requirement for slice $l$,
    \item $N_{u,m}$: number of UEs being served by RU-DU $m$.
\end{itemize}

\noindent
The objective function maximizes an overall network performance score, $\Omega(\cdot)$, while respecting slice-specific service-level constraints. Constraint~\eqref{opt1_rb} ensures resource block (RB) availability is not exceeded, and constraint~\eqref{opt1_qos} enforces minimum QoS thresholds for each slice, such as latency for URLLC or throughput for eMBB. The performance metric $Q^l_{\boldsymbol{b}, \boldsymbol{e}}$ quantifies how well the current allocation meets slice-specific service demands, and is constrained to remain above $Q^l_{\text{min}}$ to ensure acceptable service quality. 

To tackle the resource allocation problem defined in \eqref{opt1} and define an
appropriate aggregation function to represent overall network performance, we utilize the MDP framework. MDP offers a robust mathematical approach for optimizing decision-making in scenarios with inherent randomness. By modeling \eqref{opt1} as an MDP and leveraging dynamic methods like DRL, we can effectively address the complexity of this problem and achieve optimal resource allocation. 

\section{The Proposed TA-SAM MARL Network Slicing }\label{sec:SAM-MARL}
\subsection{MDP Model Representation}
The problem can be framed as an MDP, characterized by the tuples $\langle \mathcal{S}, \mathcal{A}, \mathbb{T}, \gamma, r \rangle$. In this framework, $\mathcal{S}$ denotes the state space, $\mathcal{A}$ the action space, and $\mathbb{T}$ the transition probabilities from one state to another, expressed as $P(s_{t+1}|s_t)$. Here's a detailed breakdown of each component:
\subsubsection{State} At each time step $t$, the state $s_t \in \mathcal{S}$, which is part of a continuous state space, captures the current status of the O-RAN as $s_t = \{Q^l,N^l_u,a_{t-1}\mid \forall l\in \mathcal{L}\}$. This includes the QoS for each slice ($Q^l$), the number of UEs in each slice ($N^l_u(t)$), and the previous resource allocation action ($a_{t-1}$). 
\subsubsection{Action} The action vector $a_t \in \mathcal{A}$ represents the allocation of resources to O-RAN slices and UEs as $a_t = \{\boldsymbol{e},\boldsymbol{b}\}$. At each time step $t$, the agent decides on the optimal action to take. To simplify notation, we omit the time index $t$ from variables in subsequent definitions. 
Since the chosen algorithm, SAC, inherently produces continuous-valued actions, we employ a sigmoid activation at the output of each actor network, followed by a thresholding step at inference (e.g., using 0.5 as the threshold), to convert these continuous values into discrete RB allocation decisions. 
It is important to clarify that while SAC was originally designed for continuous action spaces, in our setting the action space is a high-dimensional binary vector (twice the number of RBs, capturing both scheduling and slicing decisions). A possible alternative would be the discrete SAC formulation~\cite{christodoulou2019soft}, which models the policy as a categorical distribution using softmax. However, directly applying this approach is computationally infeasible in our case, since the joint action space grows exponentially with the number of RBs, making enumeration of all possible binary allocations impractical. 
To overcome this, we employ a continuous relaxation: the actor produces values in $(0,1)$, which are thresholded at inference to yield binary allocation decisions. This design preserves gradient-based learning in a continuous latent space while ensuring feasibility of discrete actions during execution. Although minor rounding artifacts may occur, this approach has proven effective in large-scale discrete control problems and provides a scalable, stable solution in dynamic O-RAN environments, where a categorical discrete SAC formulation would not scale efficiently.


\subsubsection{Reward} To design an efficient reward function that represents an appropriate aggregation function as overall network performance for resource utilization, we defined a utility function that integrates various parameters nonlinearly into a single reward signal. Specifically, we utilize \textit{sigmoid} functions to normalize the QoS metrics of each slice, ensuring all parameters scale uniformly between $0$ and $1$. This normalization introduces the necessary non-linearity, while accommodating diverse parameter scales effectively. Additionally, we incorporate two penalty components to address critical operational constraints, one for resource allocation and one for maintaining a minimum QoS threshold. The first penalty, $P_{\text{res}} = -\zeta \max(0,\sum_l K^l_m - K_m)$, where $\zeta$ is a scaling factor that determines the severity of the penalty and other parts are applied if the sum of resources allocated to all slices exceeds the available total, $K_m$, introducing a cost proportional to the excess used. The second penalty, $P_{\text{minQ}} = -\zeta \sum_l \max(0,Q^l_{min} - \min(Q^l))$, penalizes any instance where the minimum QoS across slices falls below a predefined threshold, $Q^l_{min}$, ensuring no slice's service quality degrades beyond acceptable limits.
The reward function at time $t$ is formulated as follows: \vspace{-0.cm}
\begin{equation}\label{reward}
r_t = \sum_{l=1}^{|\mathcal{L}|} \sigma (\alpha_l \overline{Q}^l) + P_{\text{res}} + P_{\text{minQ}} , \vspace{-0.cm}
\end{equation}
where $\sigma(x) = \frac{1}{1+ e^{-x}}$ represents the \textit{sigmoid} function, and $\overline{Q}^l$ denotes the normalized QoS for slice $l$ using min-max scaling to ensure values mapped within $[0,1]$. 
The parameter $\alpha_l$ adjusts the sensitivity of the \textit{sigmoid} function for each QoS, controlling its steepness and determining how quickly the function transitions from near $0$ to near $1$ as the input changes. To ensure a balanced reward formulation across eMBB, mMTC, and URLLC slices, we adopt an adaptive weighting approach where $\alpha_l$ is computed using a weighted normalization over $\overline{Q}^l$ with priority weights that set to $[1, 0.5, 2]$ to reflect the higher sensitivity of URLLC to latency and reliability, the moderate priority of eMBB for high throughput, and the lower sensitivity of mMTC due to its massive connectivity requirements. These adaptive coefficients $\alpha_l$ allow the DRL agent to adjust the \textit{sigmoid} sensitivity dynamically based on real-time slice conditions and preserve QoS guarantees across all slices. 
Here, we use the same penalty scaling factor $\zeta$ for both the QoS violation and RB usage constraints. This reflects the equal priority of both objectives in our O-RAN slicing formulation. Using a shared $\zeta$ also avoids introducing additional hyperparameters, simplifying training and enabling more stable convergence without manual reweighting. 

The objective is to optimize the utilization of the available bandwidth to meet the QoS requirements of all network slices. To achieve this, we model the problem as an MDP and apply a DRL approach. In the DRL framework, the agent's primary goal is to identify the optimal policy $\pi^*(a_t|s_t; \theta_p)$, where $\theta_p$ represents the parameters of the policy network. These parameters determine how the policy maps states to actions, guiding the agent's decisions. 
This policy aims to maximize the expected average discounted reward $\mathbb{E}_{\pi}[R(t)]$, where $ R(t) = \sum_{i=0}^{\infty}\gamma^i r_{i,t}.$
Here, $\gamma$ is the discount factor for future rewards. The state-value function $V(s_t)$ and the action-value function $Q_v(s_t, a_t)$ are defined respectively as $V(s_{t}) = \mathbb{E}_{\pi}[R(t)\mid s_{t}]$ and $Q_v(s_{t},a_{t}) = \mathbb{E}_{\pi}[R(t)\mid s_{t},a_{t}]$, where $\mathbb{E}_{\pi}$ denotes the expected value given the policy $\pi$.
The agent utilizes these value functions to evaluate the expected returns from different states and actions, guiding it towards the policy that maximizes long-term rewards. \vspace{-0.cm}
\begin{figure}[t!]
         \centerline{\includegraphics[width=3.5in]{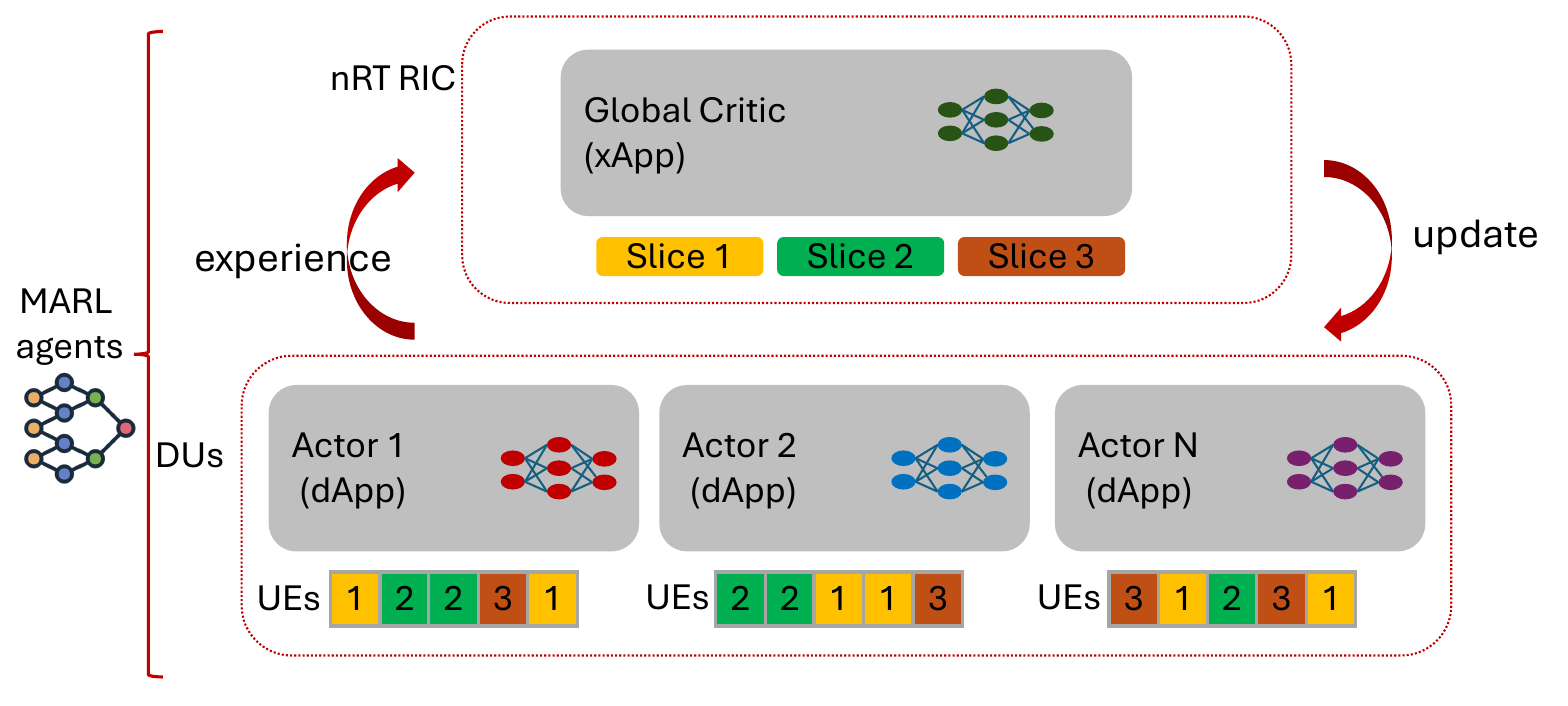}}
    \caption{\small Multi-agent RL (MARL) framework for O-RAN showing actor-critic interactions.}
    \vspace{-0cm}
    \label{marl-framework}
     \end{figure}
\subsection{MARL Framework}

Each DU is assigned a dedicated MARL agent (dApp), responsible for managing the RAN resources of its associated RUs. This design adheres to the O-RAN architectural principle of DU-level autonomy and aligns with the disaggregated control structure through the E2 interface. 
%
To build on this approach, we recognize that a single DRL agent in the near-RT RIC for O-RAN networks can be slow and suboptimal due to limited exploration. Utilizing multiple distributed DRL agents across DUs enhances training speed and stability by leveraging diverse network conditions and collective experiences. The actor-critic approach, with multiple actors and a single global critic, improves exploration-exploitation balance, generalizability, and scalability. Fig. \ref{marl-framework} represents  data flow and interaction structure in the MARL framework in O-RAN architecture. Multiple actors (dApps) handle slice-specific tasks, interact with the environment to collect experiences, and send them to the global critic (xApp) for policy evaluation and updates. The global critic aggregates experiences and provides feedback to optimize individual actor policies.
This is supported by studies that demonstrate enhanced performance in similar setups ~\cite{lotfi2022evolutionary,rezazadeh2021collaborative,kim2019reinforcement,lotfi2024open,lotfiattention,lotfi2025meta,lotfi2025oran,lotfi2025llm,lotfi2025prompt}. 
The SAC algorithm is chosen for its effectiveness in continuous environments, balancing exploration and exploitation, and achieving better sample efficiency~\cite{haarnoja2018soft}. Here, distributed intelligent actors, situated in the DU modules across the network, employ the actor-critic technique to train an optimal resource allocation policy. These actors aim to maximize long-term rewards by dynamically adapting to changing network conditions. Complementarily, a global critic, located within the CU module in the RIC 
as illustrated in Fig. \ref{sys_graph}, oversees updating the policy and value networks. It utilizes gradient and loss functions based on $\kappa$ random sample transitions to refine the policy network, parameterized by $\boldsymbol{\theta}_p$. This integrated setup ensures that the system not only optimizes resource distribution efficiently but also remains responsive to the fluctuating demands of the network environment.
\begin{align}\label{pupdate} 
    \nabla_{\theta_p}J(\pi_\theta) =  &\mathbb{E}_{\kappa,\pi}\big[ \nabla_{\theta_p} \log(\pi_{\theta_p}(a_t|s_t)) \big(-\beta \log(\pi_{\theta_p}(a_t|s_t)) \nonumber\\
    &+ Q_v(s_t,a_t;\theta_v)\big)\big],
\end{align}
where $\nabla_{\theta_p}$ represents the gradient with respect to $\theta_p$, and $\beta$ is a temperature parameter that plays a crucial role in balancing the maximization of entropy and reward within the system. Concurrently, the value network, parameterized by $\boldsymbol{\theta}_v$, is updated by minimizing the loss function defined as follows:
\begin{align}\label{vupdate} 
\min_{\theta_v} \mathbb{E}_{\kappa,\pi} \left[\left(y_t-Q_{v}(s_{t},a_{t};\theta_v)\right)^2\right],
\end{align}
where the target value $y_t$ is given by $y_t = r_t + \gamma Q_{v}(s_{t+1},a_{t+1};\theta_v) - \beta \log(\pi_{\theta_p}(a_t|s_t))$. 
Despite the success of DRL in dynamic network slicing and resource allocation, it often struggles with overfitting and adaptation to new experiences~\cite{li2017deep,zhang2018study}. To improve model generalization, the SAM optimizer is employed to find smoother minima in the loss landscape, leading to more stable and generalized learning~\cite{foret2020sharpness}. By integrating SAM into the SAC framework, we encourage convergence to these flatter minima, which indirectly assists in balancing exploration and exploitation. Flatter minima generally correspond to more stable and generalizable policies that are less prone to drastic changes in response to minor variations in input data, thus reducing overfitting and potentially enhancing exploration without compromising the exploitation of known good strategies. This is particularly beneficial in dynamic environments where continuous adaptation is crucial. Additionally, within a MARL framework, balancing exploration and exploitation using SAM enhances the DRL model's ability to efficiently adapt to ever-changing network conditions. 
While SAM has shown promise in improving generalization, its application in multi agent scenarios like MARL for dynamic network environments remains unexplored. In the studied MARL scenario, several critical questions arise: What is the optimal strategy for applying SAM in a multiple agent setting in a O-RANsystem? Given the computational cost of SAM, when should it be applied to maximize its benefits? How should the perturbation magnitude ($\rho$) be selected to balance exploration and stability in policy learning? Addressing these questions provides insights into the effective integration of SAM into MARL, shaping the core contributions of this work.





\begin{figure}[t!]
         \centerline{\includegraphics[width=2.1in]{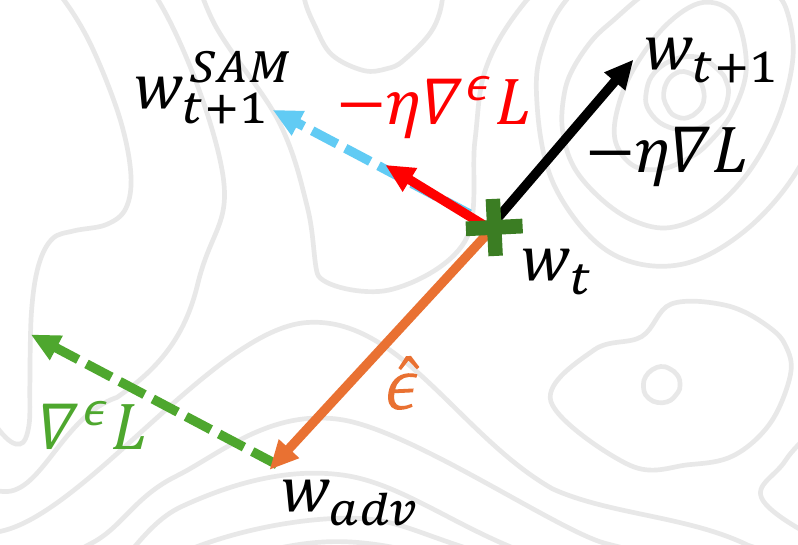}}
    \caption{\small Diagram of SAM Parameter Perturbation and Update. }\vspace{-0cm}
    \label{sam-update}
     \end{figure}
\subsection{SAM Optimizer}

To intuitively understand SAM, consider the concept of sharp versus flat minima in the loss landscape. Sharp minima typically result in high sensitivity to parameter perturbations, leading to poor generalization. Conversely, flat minima tend to yield more robust and generalizable solutions. SAM explicitly encourages convergence toward flatter regions by simulating adversarial perturbations around the current parameters. As illustrated in Fig.~\ref{sam-update}, this mechanism stabilizes learning across varying environments. 

Mathematically, SAM modifies traditional gradient descent optimization by accounting for the worst-case scenario within a specified neighborhood around the current parameters. The fundamental update rule for SAM is defined as:
\begin{equation}\label{sam_update}
    \theta_{\text{new}} = \theta - \eta \nabla_{\theta} L\left(\theta + \frac{\rho \nabla_{\theta} L(\theta)}{\left\| \nabla_{\theta} L(\theta)\right\|}\right),
\end{equation}
where $\theta$ represents the model parameters, $\eta$ is the learning rate, $L(\theta)$ is the loss function, and $\rho$ is the perturbation radius, controlling the neighborhood's extent around $\theta$. 
Specifically, the SAM optimization process involves first calculating the gradient $\nabla_{\theta} L(\theta)$, then perturbing the parameters along this gradient direction by a distance $\rho$, resulting in adversarial parameters:
\begin{align}
    \theta_{\text{adv}} = \theta + \frac{\rho \nabla_{\theta} L(\theta)}{\left\| \nabla_{\theta} L(\theta)\right\|}.
\end{align}
Next, the gradient at the perturbed parameters $\nabla_{\theta} L(\theta_{\text{adv}})$ is computed, and finally, the original parameters $\theta$ are updated using this adversarial gradient according to \eqref{sam_update}. Fig.~\ref{sam-update} illustrates this procedure.

Recent studies highlight that flat regions in the loss landscape generally correspond to improved generalization performance due to reduced sensitivity to input or training perturbations~\cite{li2018sharp}. This property is particularly critical in RL, where environment-induced variance and TD-error fluctuations can severely amplify overfitting and degrade policy performance~\cite{packer2019assessing}. Beyond these observations, \cite{achille2021information,10935668} analyze information-theoretic complexity in deep networks, providing theoretical support for variance-based measures of instability, while Optimistic Actor-Critic (OAC)~\cite{ciosek2019better} explicitly tackles exploration–stability tradeoffs in actor-critic learning.
In our setting, we denote the TD-error as $\delta_t = r_t + \gamma Q(s_{t+1},a_{t+1}) - Q(s_t,a_t)$, which serves as a stochastic sample of the Bellman residual and forms the basis for our variance-driven task complexity measure.  
Although initially introduced for supervised learning, SAM's fundamental mechanism, identifying and converging toward flatter minima, directly translates into improved stability in RL, especially within actor-critic frameworks. By reducing sensitivity to volatile TD-error gradients, SAM enhances policy robustness, resulting in higher average cumulative rewards and better generalization across dynamic and unseen network conditions, as supported by recent findings~\cite{andriushchenko2022towards,li2018sharp}. 
We adopt a dynamic scheduling of $\rho$ rather than a fixed value to ensure that sharpness-aware regularization remains responsive to evolving task complexity. Static schedules cannot adapt under varying traffic loads and slice heterogeneity, leading to either over-regularization or under-regularization in different regimes. Alternative adaptive strategies, such as entropy-based annealing, adjust only at a global level and therefore fail to capture per-task variations. In contrast, our TD-error–driven scheduling directly reflects task-specific learning difficulty, providing fine-grained adjustment of the sharpness weight. This mechanism improves responsiveness in dynamic O-RAN environments where task difficulty fluctuates rapidly across slices and users.


\subsection{Proposed SAM-Critic and SAM-Actor MARL approach}
To address the challenges of dynamic resource allocation in O-RAN, which include managing fluctuating network demands and rapidly adjusting resource distributions in real-time to maintain optimal service quality, we have developed a SAM-based MARL framework. We specifically employ SAM in the critic to stabilize learning across all tasks by smoothing the loss landscape, which is crucial in a global critic setup. In such configurations, the diversity of experiences from multiple tasks can increase the risk of erratic value estimations. By incorporating SAM, our approach mitigates these risks, ensuring consistent and reliable performance evaluations across varying network conditions, thus enhancing the efficacy of resource management in dynamic O-RAN environments. 
SAM encourages the critic to better generalize across these varied inputs, thereby reducing the variance in value estimations that could otherwise mislead actor updates. 

\begin{figure}[t!]
         \centerline{\includegraphics[width=3.5in]{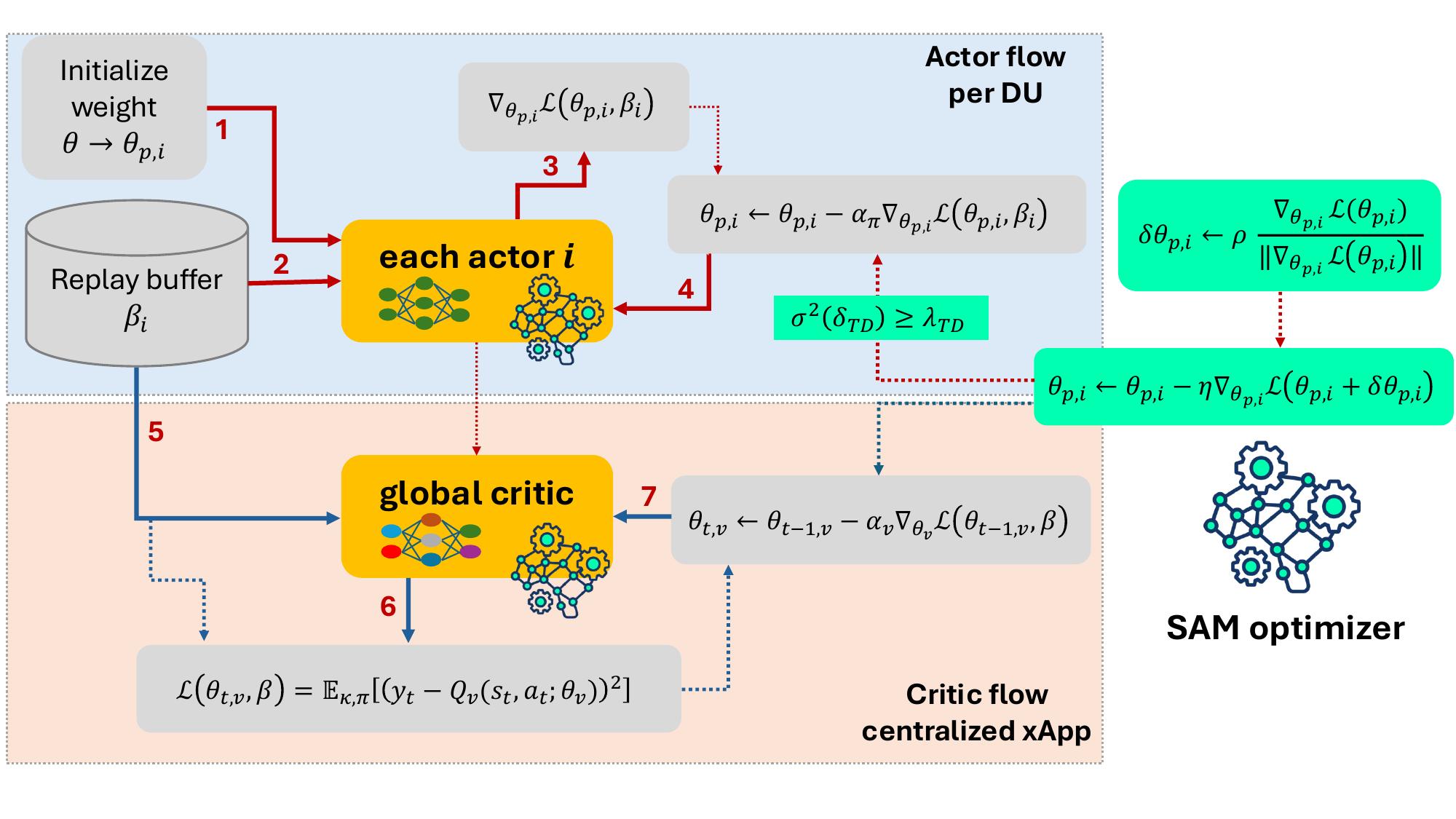}}
    \caption{\small Overview of the TA-SAM MARL training framework. The figure distinguishes actor and critic components using dashed boxes and separates their update paths through color-coded arrows. The actor flow includes: (1) actor weight initialization, (2) experience collection and replay buffer, (3) policy gradient computation, and (4) SAM-based policy parameter update. The critic flow involves: (5) accessing replay buffer data, (6) computing the critic loss, and (7) updating the critic parameters. Then, SAM perturbs policy parameters before the final update to promote flatter minima. The flow applies to all agents, while SAM updates are selectively applied only to those identified by TD-error variance. 
    } 
    \label{sam-marl-training}
     \end{figure}

Additionally, implementing SAM to the actors, especially those handling more challenging tasks, can prevent them from converging to sharp, potentially suboptimal minima. Given the increased difficulty of their tasks, these actors benefit from a more generalized policy that is robust to variations and uncertainties specific to their environments. In our proposed model, we introduce an innovative strategy to apply SAM selectively across different actor networks within a MARL framework targeting those facing complex and challenging tasks. This strategy uses the variance in the TD-error, during training, defined as  $\delta_{TD} = R_{t+1} + \gamma V(S_{t+1}) - V(S_t)$, which measures the difference between the estimated values of states and their actual outcomes. It serves as an indicator of environmental complexity and learning instability, thereby guiding the targeted deployment of SAM in areas with the most significant fluctuations. 
Recent studies have shown that TD-error variance correlates strongly with instability and task complexity in reinforcement learning. For instance, \cite{lyle2022learning} identify it as a key signal for exploration gaps and unstable value learning, while \cite{zhang2018study} associate it with poor generalization and overfitting. 
Methods such as TD-regularized actor-critic~\cite{parisi2019td} further demonstrate the benefit of stabilizing updates by penalizing high TD errors in the learning objective. In curriculum learning contexts, TD-error guides sampling or regularization to focus resources on high-difficulty states~\cite{banerjee2023boosting,xu2024vertiselector}. From a theoretical perspective, the TD error is a stochastic sample of the Bellman residual, and classical error-propagation analyses show that performance degradation is bounded by norms of this residual~\cite{munos2003error,munos2007performance,farahmand2010error}. Consequently, high TD-error variance naturally reflects regions of greater approximation difficulty and instability. 
Motivated by these findings, we use TD-error variance as a dynamic signal to selectively apply SAM regularization to actor networks operating in more complex regions of the environment. This strategy enhances the generalization capability of the actors by smoothing their loss surfaces only when instability is detected, improving convergence stability while avoiding unnecessary regularization overhead. 
By smoothing the loss landscape, SAM prevents actors from converging to sharp, potentially suboptimal minima, enhancing policy robustness and generalization. Fig.~\ref{sam-marl-training} represents this process.  

The effectiveness of SAM in our MARL framework depends significantly on the appropriate tuning of its hyperparameter, $\rho$, which dictates the magnitude of the perturbations applied during optimization. This parameter may require different settings for different actors within the same system, depending on the complexity and specific challenges of the tasks they face. Properly adjusting $\rho$ ensures that each actor is neither too sensitive to small changes in the input, which can lead to instability, nor too insensitive, which might prevent effective learning and adaptation to new or complex scenarios.

\subsection{Adjusting hyperparameter $\rho$}

In deploying SAM within DRL frameworks like SAC, the hyperparameter $\rho$ is pivotal, representing the size of the neighborhood considered around the current parameters to simulate adversarial conditions. The careful adjustment of $\rho$ plays a critical role in fine-tuning the exploration-exploitation balance. This section delves into how variations in $\rho$ affect overall system performance, illustrating the necessity of its dynamic adjustment to enhance the generalization and responsiveness of the learning models across different network environments. Specifically, $\rho$ determines the radius of the perturbation applied to the model parameters during optimization, dictating the extent of the neighborhood around the current parameters considered for minimizing the loss. A higher $\rho$ introduces larger perturbations, steering the optimizer to explore a broader area of the parameter space. This broader consideration helps the model avoid settling in sharp minima, which are typically sensitive to input variations and prone to overfitting, thereby enhancing exploration. Conversely, a lower $\rho$ leads to smaller perturbations, focusing the optimization on a tighter region around the current estimates, which enhances exploitation by fine-tuning the model within a more confined scope. Properly balancing $\rho$ is thus essential for achieving an effective trade-off between exploring potential new strategies and exploiting known rewarding strategies. Too high a $\rho$ can cause the model to over-explore without effectively learning optimal behaviors, while too low a $\rho$ might restrict the learning to suboptimal policies. In practice, dynamically adjusting $\rho$, starting higher for initial exploration and reducing it as the model's performance stabilizes, can optimize learning outcomes in complex environments.

Recent studies have explored how entropy regularization can smooth the optimization landscape in RL, leading to improved generalization performance \cite{ahmed2019understanding}. Additionally, techniques like SAM have been proposed to find flatter minima, which are associated with better generalization in deep learning models \cite{foret2020sharpness,andriushchenko2022towards}. These findings provide further theoretical motivation for integrating curvature-aware strategies like SAM into RL frameworks such as ours. 
While entropy-regularized methods like SAC inherently manage exploration via policy entropy, our dynamic $\rho$ adjustment provides gradient-level control over update smoothness, orthogonal to entropy-based methods. The two mechanisms act on different aspects of the optimization process: entropy promotes stochastic policy behavior, whereas SAM reduces sensitivity to local perturbations. This orthogonality allows their combination to yield more generalizable and stable policies in non-stationary O-RAN environments.
Hence, we use dynamic scheduling of $\rho$ where the value decreases over training time, starting with a higher value to encourage exploration in the early stages of training and reducing it gradually to promote exploitation of the learned policy as the agent matures. 
\begin{algorithm}[t!]
\SetAlgoLined
\textbf{Input}: Number of iterations $N_t$, actors $N_m$, evaluations $N_e$, actor weights $\theta_{p,i}$, critic weights $\theta_{v}$.  \\
\textbf{Initialize:} Randomly initialize $\theta_{p,i}$, $\theta_{v}$. \\
\For{iteration $t=1:N_t$}{
\For{actor $i=1:N_m$}{
Collect experiences and evaluate policy $\pi_{p,i}$ to get rewards.\\
\For{evaluation $e = 1 : N_e$}{
$r_i = \text{evaluate}(\pi_{p,i})$.\\
Store experiences in buffer $\mathcal{B}_i$\\
$\mathcal{B}\gets \langle s_t,a_t,s_{t+1},r_t \rangle $.
}
\eIf{$\sigma^2(\delta_{TD})\geq \lambda_{TD}$}{
Apply SAM optimizer on actor. \\
Compute perturbation as $\delta \theta_{p,i} \gets \rho_{a,t} \cdot \frac{\nabla_{\theta_{p,i}} L(\theta_{p,i})}{\left\|\nabla_{\theta_{p,i}} L(\theta_{p,i})\right\|}$.\\
Update weights by \eqref{sam_update} as $\theta_{p,i} \gets \theta_{p,i} - \eta \cdot \nabla_{\theta_{p,i}} L(\theta_{p,i} + \delta \theta_{p,i})$}
{
 Apply ADAM optimizer on actor. \\
 Update weights by SGD.
 }
 }
Global Critic update.\\
Compile all experiences into $\mathcal{B}$ and apply SAM on Critic.\\
Compute perturbation as $\delta \theta_{v} \gets \rho_{c,t} \cdot \frac{\nabla_{\theta_{v}} L(\theta_{v})}{\left\|\nabla_{\theta_{v}} L(\theta_{v})\right\|}$.\\
Update weights by \eqref{sam_update} as $\theta_{v} \gets \theta_{v} - \eta \cdot \nabla_{\theta_{v}} L(\theta_{v} + \delta \theta_{v})$.\\
 \If{$\theta_p$ by \eqref{pupdate} is converged}{
 Break.
 }
}
\textbf{Output}: Trained policy weights $\theta_{p,i}$ for each actor, updated critic weights $\theta_{v}$.\\
\caption{TA-SAM MARL algorithm}\vspace{-0cm}
\label{alg1}
\end{algorithm}\vspace{-0.cm}

\subsection{Proposed TA-SAM MARL Algorithm}
A DRL agent learns to efficiently allocate shared resources across different network slices, aiming to optimize overall network efficiency and meet global QoS metrics. The integration and coordination of the SAM enhanced MARL for network slicing and resource allocation are systematically outlined in Algorithm~\ref{alg1}, which addresses the optimization problem as defined in equations~\eqref{opt1}-\eqref{opt1_qos}. 
The algorithm initializes with input variables including the total number of training iterations $N_t$, the number of distributed actors $N_m$, the frequency of evaluations $N_e$, and the initial random weights for the distributed actors’ network weights $\theta_{p,i}$ and the critic network weights $\theta_v$. Each training iteration involves $N_e$ evaluations per distributed actor, allowing sufficient interactions with the environment. Experiences from these interactions are stored in the replay buffer $\mathcal{B}$. Subsequently, the critic network updates $\theta_v$ based on \eqref{vupdate}, and in turn updates the distributed actors’ networks using \eqref{pupdate}. In our model, SAM optimizers are employed in both critic and actor updates. For actors, SAM is applied dynamically when the variance of the TD-error, $\sigma^2(\delta_{TD})$, exceeds a predefined threshold $\lambda_{TD}$, ensuring stability and robustness in more complex scenarios. Additionally, the $\rho_{a,t}$ and $\rho_{c,t}$ parameters are adjusted dynamically, decreasing over the course of training $t$ to smoothly transition from a focus on exploration to exploitation. This adaptive approach allows for more effective learning and generalization as the environment's challenges evolve. 
The process iterates until the policy networks of the distributed DRL agents have converged or the maximum iteration limit $N_t$ is reached.\vspace{-0.0cm}
\begin{table}[t!] 
	\footnotesize
	\centering
	\caption{\vspace*{-0cm} Simulation parameters} \vspace{-0.cm}
	\begin{tabular}{|>{\centering\arraybackslash}m{2.4cm}|>{\centering\arraybackslash}m{1.7cm}|>{\centering\arraybackslash}m{1.2cm}|>{\centering\arraybackslash}m{1.7cm}|}
		\hline
		\bf{Parameter} &\bf{Value } & \bf{Parameter} &\bf{Value }\\
		\hline
		Subcarrier spacing & $15$ kHz & $h$ & Rayleigh fading channel \\
		\hline
		Total bandwidth & $20$ MHz  & $\sigma^2$ & $-173$ dBm \\
		\hline
		RB bandwidth  & $200$ kHz & $N_t$ & $1000$ \\
		\hline
		$p_u$ & $56$ dBm & $N_m$ DU & $6$ \\
		\hline
		$K$ RBs& $100$ & $N_e$ & $10$ \\
		\hline
		$N$ UEs & $200$ & batch size & $128$ \\
		\hline
	\end{tabular}\label{param} \vspace{-0.cm}
\end{table}

\section{Proposed Algorithm Analysis}
This section analyzes various aspects of the proposed TA-SAM MARL algorithm.

\subsection{Algorithm Training Framework } 
\textbf{Multi Agent Learning: }The training framework consists of multiple distributed agents (actors), each representing a Distributed Unit (DU) and a global critic (xApp) housed within the Near-RT RIC. The actors interact with the O-RAN environment by making resource allocation decisions, while the global critic evaluates and refines their policies based on shared experiences.
The SAC algorithm is employed to prevent instability in learning and poor generalization. SAC effectively balances exploration and exploitation using entropy regularization, making it well-suited for dynamic network slicing scenarios.

\textbf{Selective SAM: } One of the main challenges in multi-agent learning is ensuring that policies generalize across varying network conditions. To address this, SAM is selectively applied to actor networks that exhibit high environmental complexity, identified through variance in TD-error. This adaptive application of SAM prevents sharp minima and improves the stability and robustness of learned policies.
Unlike conventional regularization methods, SAM directly influences the loss landscape, encouraging flatter minima. This approach ensures that policies remain generalizable and do not overfit to transient network conditions.

\textbf{Data Flow between Actors and Global critic: }
Actors (i.e., DUs) collect state-action-reward transitions and send them to the global critic. Then, the global critic aggregates experiences, evaluates policy performance, and computes gradient updates. Next, the calculated policy updates are returned to the actors, refining their resource allocation strategies. This decentralized actor-critic architecture ensures that local agents (DUs) adapt to their respective network slices while benefiting from a centralized learning signal.

\textbf{Adaptive Parameter Adjustment: } 
One of the standout features of the TA-SAM MARL framework is its dynamic adjustment of the $\rho$ parameter, which helps balance exploration and exploitation. Instead of using a fixed value, $\rho$ adapts to the complexity of the environment, allowing the learning process to stay responsive to real-time changes in network conditions. 
We adopt a dynamic scheduling of $\rho$ rather than a fixed value to ensure that sharpness-aware regularization remains responsive to evolving task complexity. Static schedules cannot adapt under varying traffic loads and slice heterogeneity, leading to either over-regularization or under-regularization in different regimes. Alternative adaptive strategies, such as entropy-based annealing, adjust only at a global level and therefore fail to capture per-task variations. In contrast, our TD-error driven scheduling directly reflects task specific learning difficulty, providing fine grained adjustment of the sharpness weight. This mechanism improves responsiveness in dynamic O-RAN environments where task difficulty fluctuates rapidly across slices and users. 
This adaptability improves learning efficiency and ensures stable convergence, making the model more effective in dynamic settings.


\subsection{Computational Complexity and Scalability of Algorithm}
The computational complexity of the proposed TA-SAM MARL algorithm primarily stems from the SAC updates, SAM optimization, and the distributed multi-agent architecture. Specifically, SAC entails actor-critic updates and entropy regularization, incurring a per-step complexity of $\mathcal{O}(N_m \cdot \hat{d}_{s,a})$, where $N_m$ is the number of agents (DUs), and $\hat{d}_{s,a}$ denotes the state-action dimensionality. 
SAM introduces additional overhead due to its two-step gradient update mechanism, requiring an extra forward and backward pass per parameter update. While this increases per-update cost, the resulting flatter optima significantly enhance model generalization and sample efficiency, reducing long-term training cycles. 
In terms of scalability, our decentralized actor–centralized critic structure enables parallel training of agents, minimizing inter-agent communication overhead. Moreover, the adaptive $\rho$ tuning mechanism ensures computational efficiency by selectively applying SAM to agents/slices that benefit most from sharper minimization.

\textbf{Comparison with PLASTIC Baseline:} 
To contextualize the efficiency of our method, we compare its computational aspects with PLASTIC~\cite{lee2024plastic}, a state-of-the-art MARL distillation framework. PLASTIC relies heavily on knowledge distillation from teacher to student agents, incorporating both input-plasticity and label-plasticity losses. These additional components, particularly the ensemble based supervision and input perturbation modules, introduce overhead in both forward passes and memory usage due to storing multiple teacher signals and augmented input trajectories. By contrast, our TA-SAM MARL approach avoids reliance on teacher-student transfer, maintaining a simpler training pipeline and reducing architectural complexity. Moreover, while both approaches incur extra computation for better generalization, our method uses selective SAM scheduling to dynamically modulate this cost, offering more control over computational overhead. Although PLASTIC improves sample efficiency through distillation, our approach strikes a better balance between complexity, stability, and generalization via SAM-based regularization, and is more suitable for distributed RAN systems with decentralized decision-making needs. 

\subsection{Catastrophic Forgetting During Agent Training}
One of the challenges in DRL for dynamic environments like O-RAN is catastrophic forgetting, where an agent fails to retain valuable knowledge from past interactions as it continually updates its policy in response to new experiences. This issue is particularly problematic in multi agent settings where distributed agents operate in a constantly changing environment, requiring frequent policy adaptations. Unlike static tasks, where training data distribution remains stable, O-RAN environments exhibit non-stationary data distributions, as network slices experience fluctuations in user density, traffic demand, and QoS requirements. As a result, traditional RL methods, particularly off-policy algorithms like SAC, tend to overwrite previously learned policies when optimizing for new observations, leading to performance degradation over time.
In our proposed SAM-enhanced MARL framework, we address catastrophic forgetting through two key mechanisms as gradient smoothing via SAM and experience replay with stability constraints. The integration of SAM in both the actor and critic networks mitigates abrupt policy shifts by minimizing sharp variations in the loss landscape, encouraging convergence to flatter minima that promote generalizability. This ensures that learned policies remain robust across varying network conditions rather than being optimized solely for recent experiences. Furthermore, SAM enhances the stability of temporal difference learning by preventing over-aggressive updates in regions where TD-error variance is high, which is a primary contributor to catastrophic forgetting in MARL.

To further reinforce knowledge retention, our method leverages a distributed experience replay mechanism, where global and local critics aggregate past experiences from multiple agents, maintaining historical information alongside new updates. Instead of letting each agent completely overwrite past policies, we use a dynamic regularization mechanism. It penalizes drastic changes from previously successful policies, helping the agent adapt while preserving important knowledge. Additionally, our framework dynamically tunes the SAM perturbation factor $\rho$ based on TD-error variance, adapting its influence to prevent catastrophic forgetting in high-variance regions while enabling sufficient policy refinement in stable environments.
Our approach combines gradient smoothing, distributed experience aggregation, and adaptive regularization to prevent catastrophic forgetting. This ensures stable performance over time and keeps O-RAN network slicing and scheduling reliable. 

\begin{figure*}
     \centering
     \begin{subfigure}[a]{0.22\textheight}
         \centerline{\includegraphics[width=2.5in]{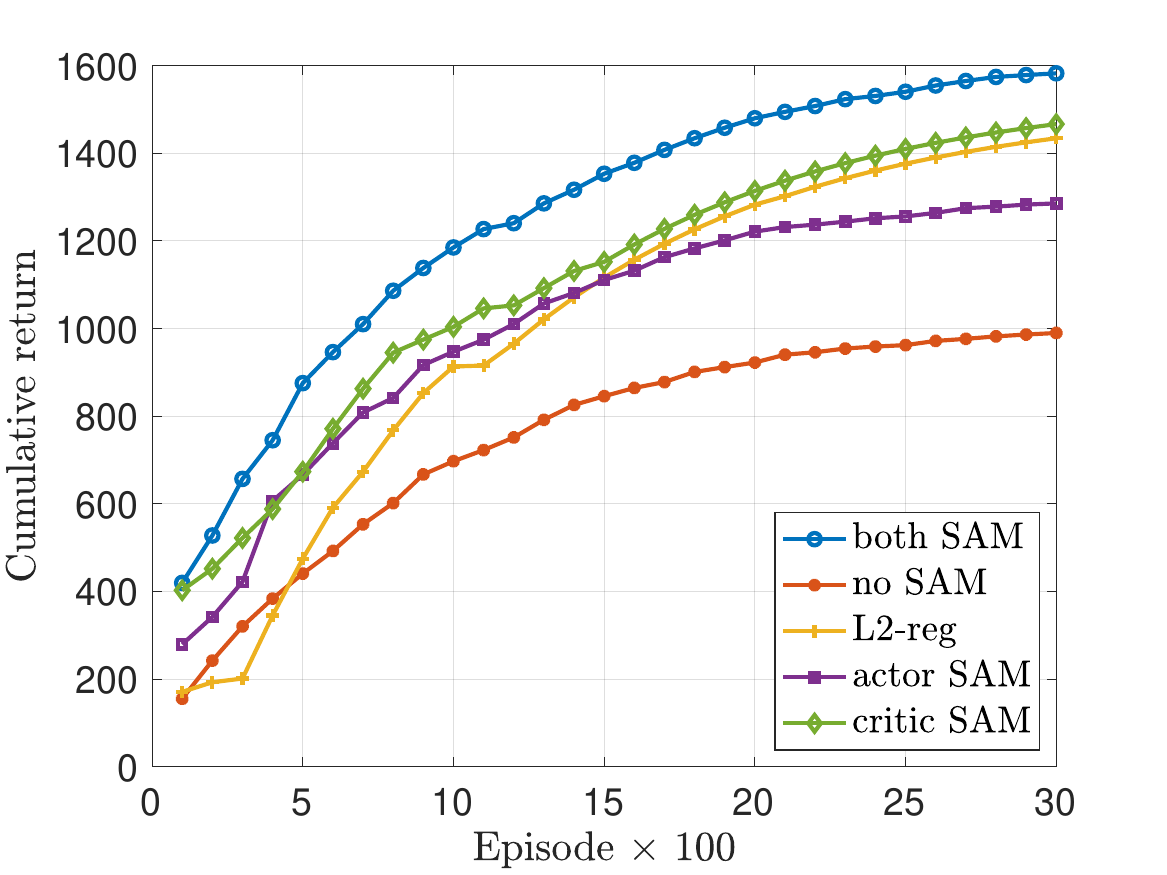}}
    \caption{\small equal $\rho$ scenario, $\rho_{actor} = \rho_{critic}=0.01$.  (sample)
    }\vspace{-0cm}
    \label{rew_rho_eq}
     \end{subfigure}
     \hfill
     \begin{subfigure}[a]{0.22\textheight}
         \centerline{\includegraphics[width=2.5in]{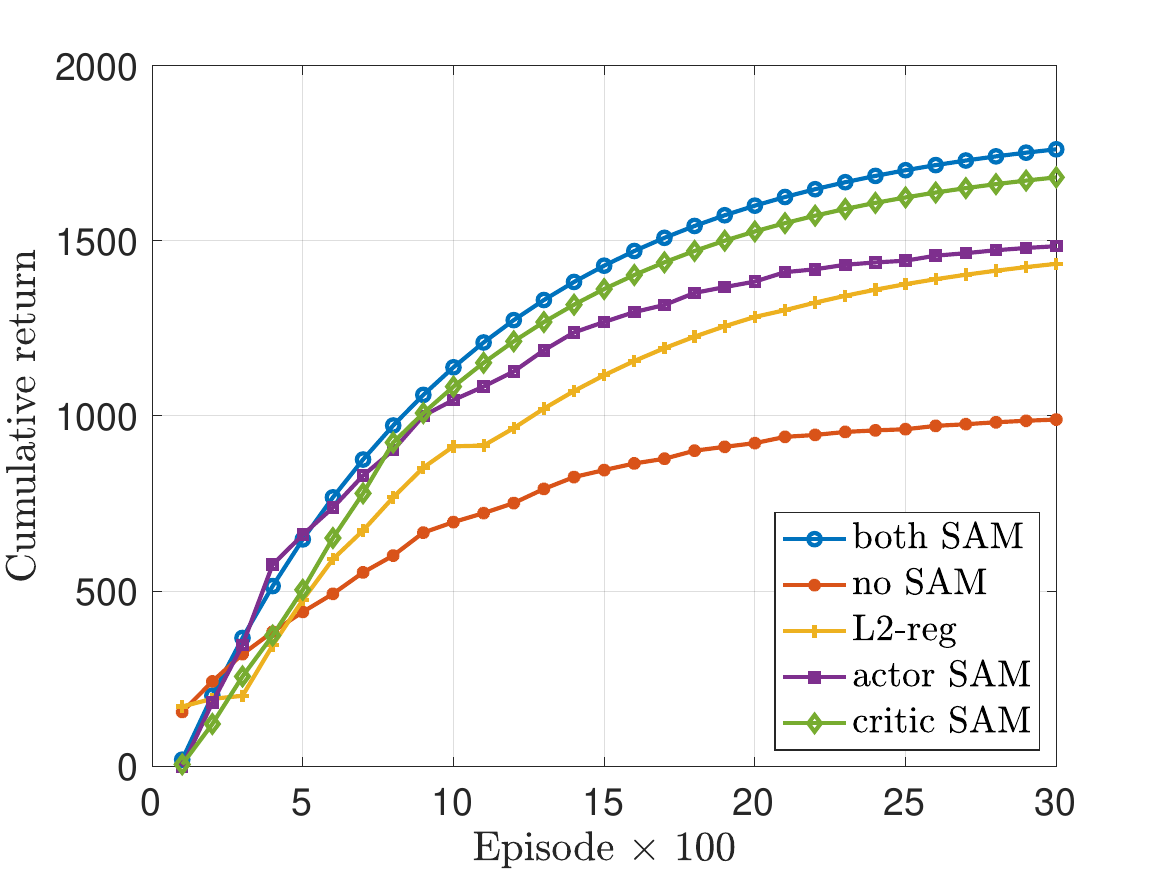}}
    \caption{\small non-equal $\rho$ scenario, $\rho_{actor}=0.05$ and $\rho_{critic}=0.01$.  (sample)
    }\vspace{-0cm}
    \label{rew_rho_neq}
     \end{subfigure}
     \hfill
     \begin{subfigure}[a]{0.22\textheight}
         \centerline{\includegraphics[width=2.5in]{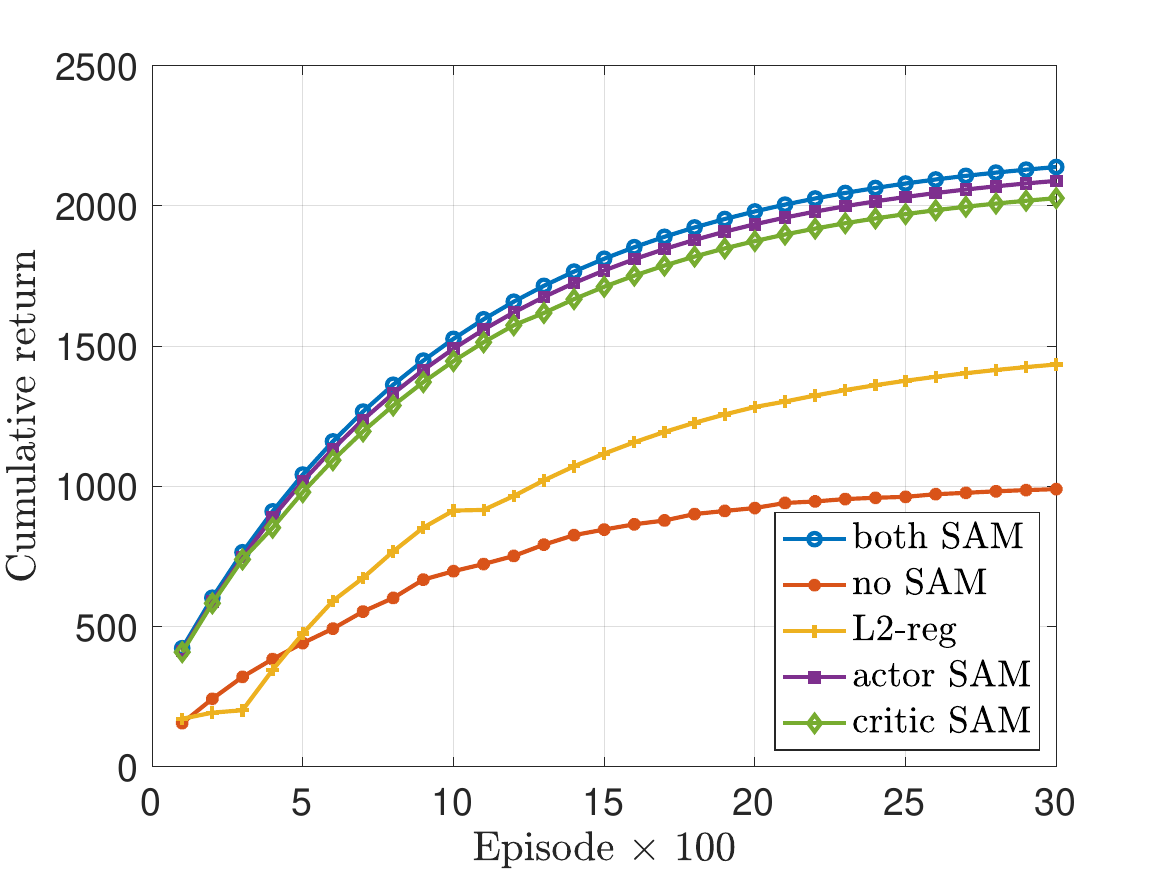}}
    \caption{\small dynamic $\rho$ scenario, uniformly from $0.5$ to $0.01$.
    }\vspace{-0cm}
    \label{rew_rho_adp}
     \end{subfigure}
        \caption{Average cumulative reward values in different $\rho$ selection scenarios} 
        \label{rew_rho}
        \vspace{-0.cm}
\end{figure*}

\section{Evaluation Results}\label{sec:simulation}

\subsection{Simulator and Parameter Settings}
In our O-RAN simulation framework, we utilize an architecture configured to support three network slices, eMBB, MTC, and URLLC, servicing a total of $200$ users, which are distributed according to their service demands and uniformly spread across the network. Our setup includes $N_m = 6$ DUs, each covering different network areas and acting as distinct agent actors. This configuration addresses the heterogeneous distribution of demand services, with each DU encountering varied levels of UE traffic and requiring specific service prioritization. The RB bandwidth is set to $200$ kHz, approximating the $180$ kHz RB width in 3GPP NR numerology $\mu=0$, with additional buffer to account for guard bands and control overhead, as common in software-defined RAN simulators. 
User mobility is modeled with speeds ranging between $10 m/s$ and $20 m/s$, moving in one of seven possible directions $\{\pm \pi/3, \pm \pi/6, \pm \pi/12, 0\}$, across DU-designated areas. The traffic dynamics, influenced by UE density and mobility within the cell range, are developed following the methodologies described in~\cite{cheng2022reinforcement, lotfi2024open}. To implement our TA-SAM MARL strategy, we adopt an actor-critic method using the Pytorch library, configured with three fully-connected layers containing $300$, $400$, and $400$ neurons for both the actor and critic networks, using a \textit{tanh} activation function. All models utilize a learning rate of $10^{-4}$ and the \textit{Adam} optimizer.

We evaluate our TA-SAM MARL approach across several scenarios, integrating SAM selectively in actor networks, critic networks, both, or not at all, and compare these to a baseline SAC model employing L2 regularization. L2 regularization is used as a baseline because it is a standard method to prevent overfitting and enhance model generalization by limiting the magnitude of network parameters. This comparative analysis not only demonstrates the efficiency and effectiveness of our proposed approach but also highlights its advantages over traditional RL strategies that rely on simpler regularization techniques. Additional details on other parameters are provided in Table~\ref{param}.\vspace{-0.cm}

\subsection{Cumulative reward of TA-SAM MARL}
This section explores proposed approach learning curve and tracks the cumulative reward (or return) collected by the agent over episodes or time steps to demonstrates how the agent learns to maximize rewards as training progresses. Moreover, graphs explore the impact of various approaches on cumulative rewards, with a focus on the influence of $\rho$ selection within the TA-SAM MARL framework. 
Fig.~\ref{rew_rho_eq}, Fig.~\ref{rew_rho_neq}, and Fig.~\ref{rew_rho_adp} show the average cumulative return graphs for the scenarios with equal $\rho$, non-equal $\rho$, and dynamic $\rho$ settings across different SAM optimizer configurations in actor and critic networks and reveals key insights into their performance. In the equal $\rho$ setting Fig. \ref{rew_rho_eq}, where the same $\rho$ value is used for both actor and critic SAM optimizers, the \textit{both-SAM} method consistently delivers superior performance, achieving the highest cumulative returns. This indicates a well-balanced $\rho$ optimization that effectively enhances overall system performance. 

\begin{figure}[t!]
    \centerline{\includegraphics[width=3.5in]{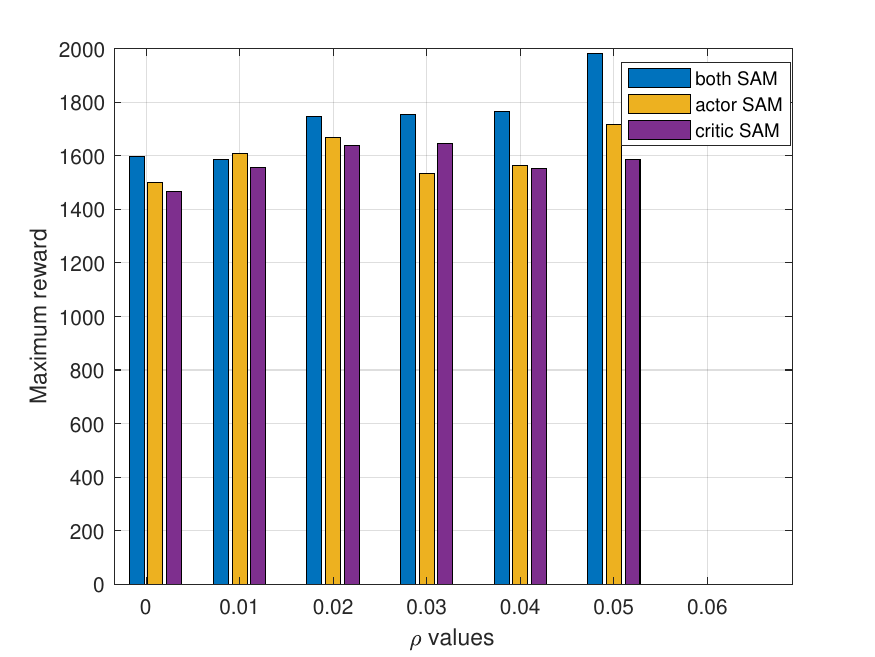}}
    \caption{ Comparison of the impact of $\rho$ values on maximum cumulative reward in scenarios with equal $\rho$ settings. 
    }\vspace{-0cm}
    \label{comp_rho}
    \vspace{-0.cm}
\end{figure}

Conversely, in the non-equal $\rho$ scenario Fig. \ref{rew_rho_neq}, where $\rho$ values are non-equal between actor and critic, \textit{both-SAM} shows better performance, suggesting that fine-tuning $\rho$ values individually for different network components can optimize specific performance metrics. In the dynamic $\rho$ scenario Fig. \ref{rew_rho_adp}, where $\rho$ values are dynamically adjusted during training, the methods performance are different, with \textit{both-SAM} reaching a higher gain in comparison other methods. This approach demonstrates the potential to maximize performance by continuously recalibrating the optimizer parameters in response to training dynamics, effectively smoothing out disparities seen in more static $\rho$ configurations. Moreover, Fig~\ref{rew_rho} demonstrates that our approach attains higher cumulative rewards more quickly than traditional methods, necessitating fewer iterations for comparable results. On the other hand, Fig. \ref{comp_rho} illustrates the maximum reward achieved under varying $\rho$ values for different SAM configurations methods. The graph demonstrates how each $\rho$ setting impacts the system's performance, indicating the $\rho$ values that optimize gains across scenarios.

\begin{figure}[t!]
    \centerline{\includegraphics[width=3.5in]{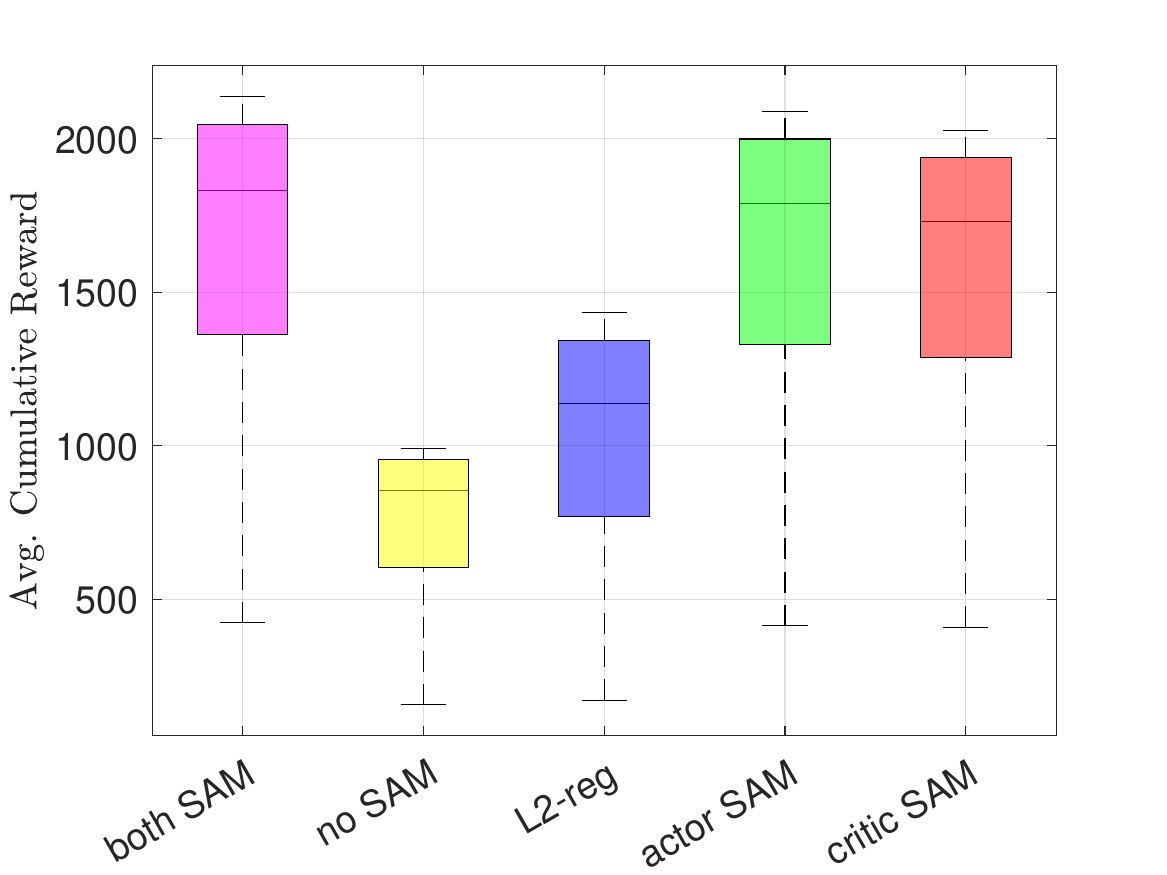}}
    \caption{ Distribution of average cumulative reward across various approaches in a dynamic $\rho$ scenario. 
    }\vspace{-0.cm}
    \label{dist_rho}
    \vspace{-0.cm}
\end{figure}

Fig. \ref{dist_rho} illustrates the distribution of average cumulative rewards across different approaches in a dynamic $\rho$ setup including \textit{no-SAM}, \textit{L2-reg}, \textit{actor-SAM}, \textit{critic-SAM}, and \textit{both-SAM}. This comparison shows that applying SAM, whether to the actor, the critic, or both, generally enhances performance over the traditional SAC approach without SAM (\textit{no-SAM}). Notably, \textit{both-SAM} achieves the highest median cumulative reward, indicating a synergistic effect when SAM is applied comprehensively. The \textit{critic-SAM} configuration slightly outperforms \textit{actor-SAM}, suggesting more effective use of SAM in the critic role for this scenario. However, \textit{L2-reg}, while improving over \textit{no-SAM}, does not reach the performance heights of SAM-enhanced configurations. The interquartile ranges (IQR), which represent the middle $50\%$ of the data, demonstrate performance variability, with \textit{both-SAM} exhibiting the least variability and indicating consistent performance across various trials. 
Based on the results in Fig. 5, we adopt a dynamic selection strategy for $\rho$. Initially, a larger $\rho$ value is chosen to enable the model to explore a broader region of the loss landscape to enhance flatness and improve generalization. As training progresses, $\rho$ is gradually reduced from 0.5 to 0.01  shifting the focus toward fine-tuning near optimal values rather than maintaining a flattened loss surface. This adaptive approach balances exploration and convergence, leading to more stable and effective optimization.

\subsection{Loss Curvature and Generalization Analysis} 
To evaluate the generalization performance of different approaches, we analyze the curvature of the loss landscape by computing the maximum eigenvalue of the Hessian ($\lambda_{max} (\nabla^2 \mathcal{L}) $) as shown in \cite{lee2024plastic}. The Hessian encapsulates the second-order derivatives and shows how rapidly the loss changes with respect to model parameters. A larger maximum eigenvalue indicates a sharper loss landscape, often correlating with poorer generalization, and a smaller maximum eigenvalue suggests a flatter region, where the model is more resistant to overfitting and better performance at generalizing to unseen data. By comparing eigenvalues across different regularization techniques, we can quantitatively assess which approach finds flatter minima. Fig.~\ref{max_eig} illustrates the curvature of the loss landscape by showing the maximum eigenvalue of the Hessian matrix across different approaches. 
This graph shows effectiveness of SAM optimizer in reducing loss landscape curvature. 



\subsection{Scalability with Number of Agents}


\begin{figure}[ht]
    \centering
    \includegraphics[width=0.45\textwidth]{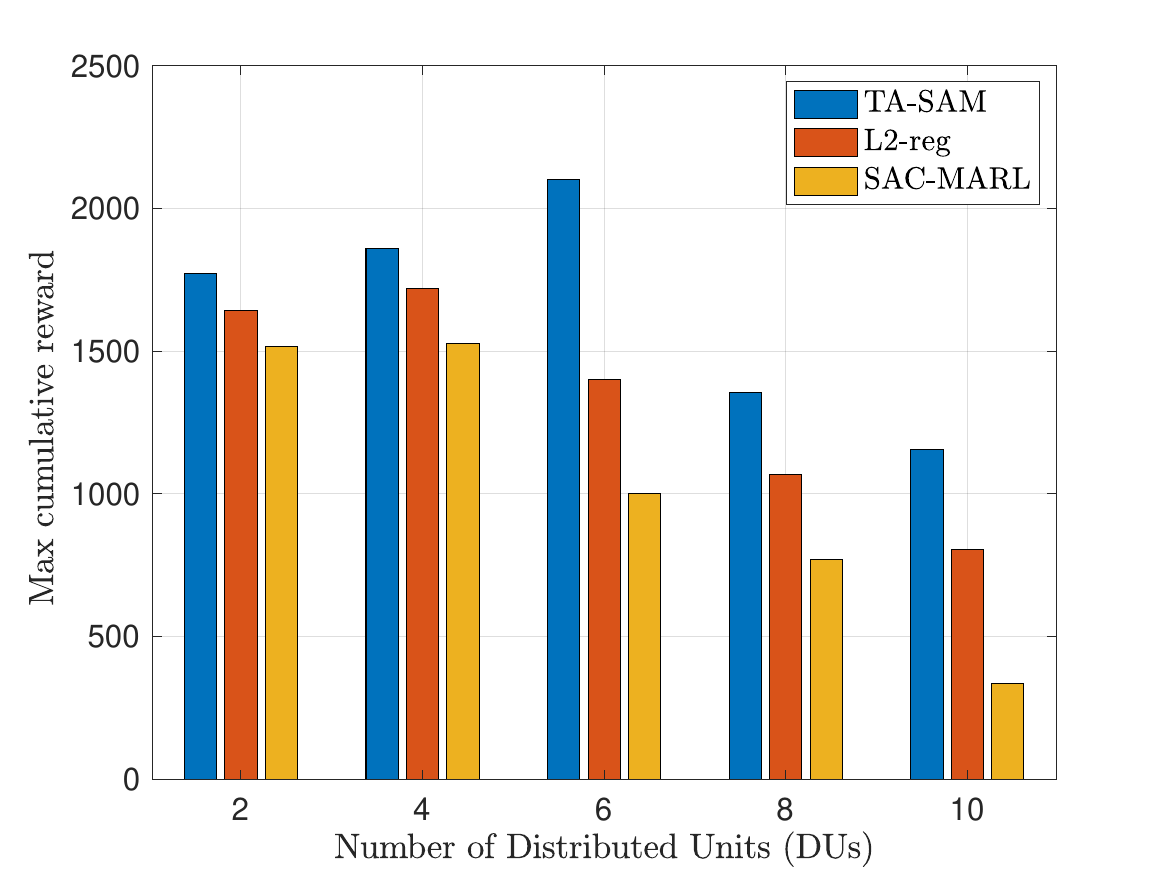}
    \caption{ Scalability analysis of cumulative reward with increasing DUs. 
}
    \label{fig:scalability}
\end{figure}

Figure~\ref{fig:scalability} presents the sensitivity of cumulative reward as the number of DUs increases from 2 to 10. As the result shows, performance initially improves up to 6 DUs, likely due to more effective distributed learning and resource utilization. Beyond this point, performance begins to decline, driven by increased coordination overhead, communication latency, and limited resource availability. Despite this trend, TA-SAM maintains a clear advantage over other baselines like SAC-MARL and L2-regularization across all scales, demonstrating its robustness in handling system complexity.
\subsection{QoS in Network Slices}\vspace{-0.cm}
In our evaluation study, we consider QoS metrics according to the specific requirements of each network slice, thereby optimizing the performance and service delivery for diverse user demands. For the eMBB as slice $1$, we focus on the aggregated network user throughput, which reflects the high-throughput demands typical of enhanced mobile broadband services. For the mMTC as slice $2$, we employ a novel metric of mix of multiple KPIs as the Quality-Weighted Network Capacity, which combines the aggregated users throughput with a coefficient representing the proportion of users meeting a satisfactory service threshold as UE density support. This metric is particularly effective in measuring mMTC performance as it balances the need for widespread connectivity with the quality of service experienced by individual devices. Lastly, for the URLLC as slice $3$, we consider the maximum user transmission latency as the QoS metric, prioritizing the ultra-reliability and low latency required by applications such as real-time control systems and emergency services. 
\begin{figure}[t!]
    \centerline{\includegraphics[width=3.5in]{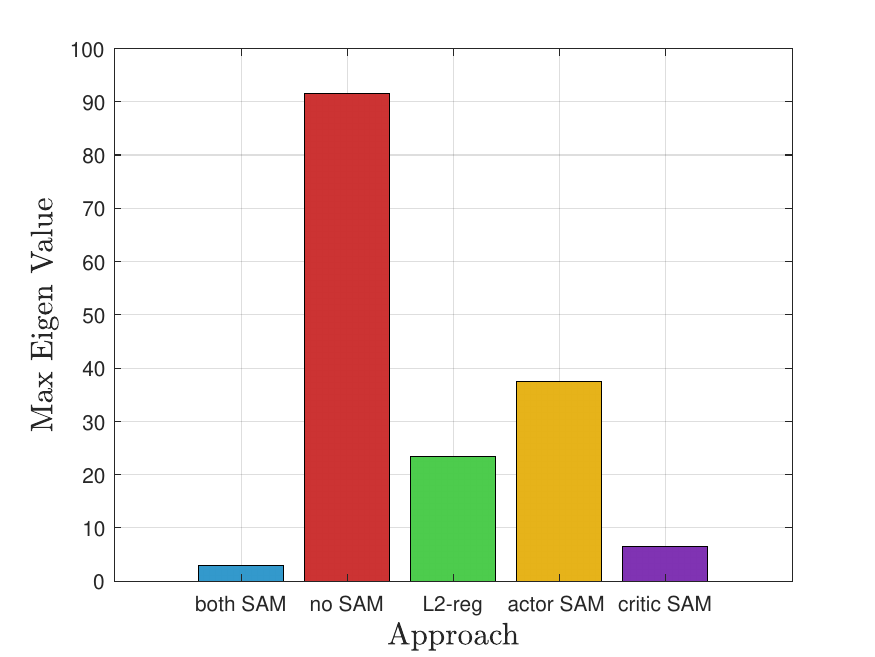}}
    \caption{ Representation of the loss landscape curvature through the maximum eigenvalue of the Hessian matrix (standard measure of sharpness)}. 
    \vspace{-0cm}
    \label{max_eig}
    \vspace{-0.cm}
\end{figure}

\begin{figure}[t!]
    \centerline{\includegraphics[width=3.5in]{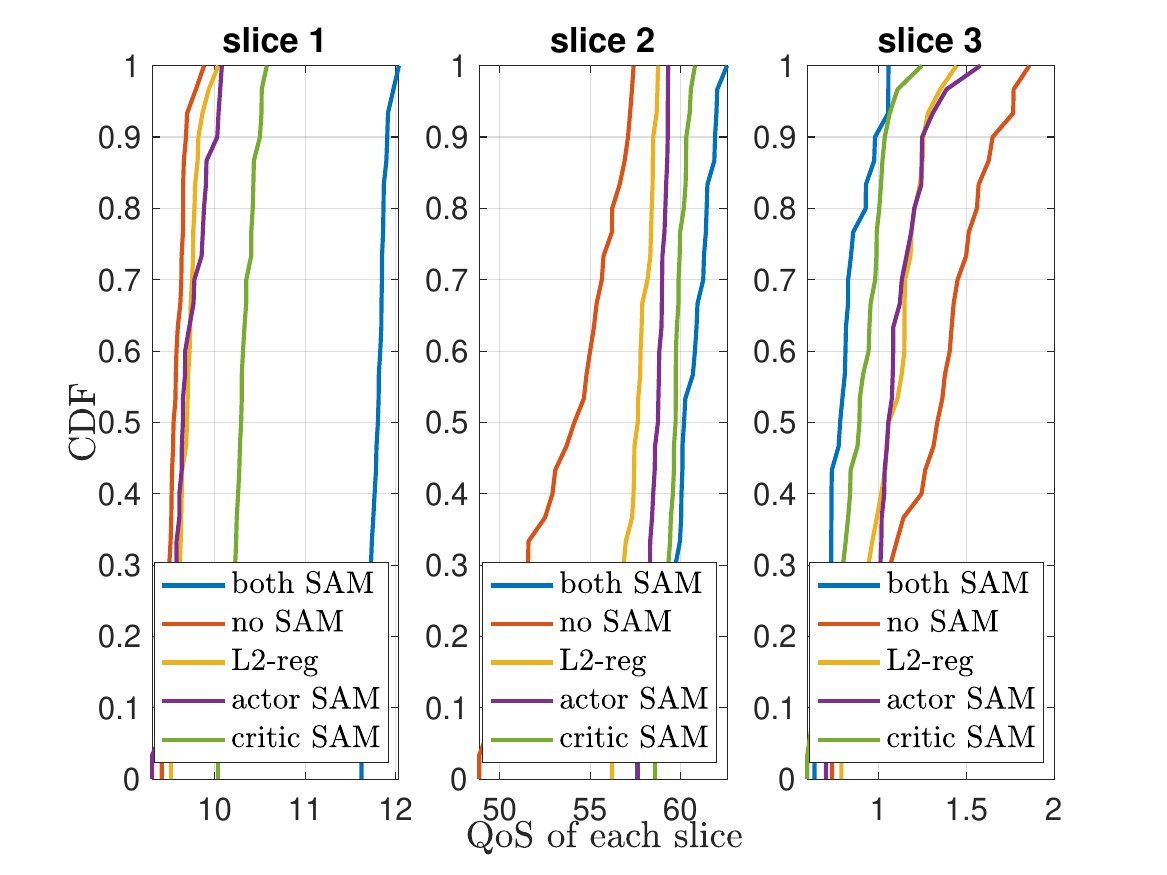}}
    \caption{ Comparison of network slices' QoS CDF across different approaches in a dynamic $\rho$ scenario.
    }\vspace{-0cm}
    \label{slice-Qos}
\end{figure}

Fig. \ref{slice-Qos} displays the Cumulative Distribution Function (CDF) of QoS for different network slices, highlighting the effectiveness of various methods in achieving maximum QoS according to the unique requirements of each slice. The graph indicates that the \textit{both-SAM} method excels by rapidly achieving higher throughput in slices $1$ and $2$ and maintaining low latency in slice $3$, quickly surpassing the $Q^l_{min}$ of $10$ Mbps, $50$ Mbps, and $2$ ms, respectively~\cite{embb_urllc_mtc_thr}. This method's CDF curves rise rapidly, demonstrating its superior ability to effectively meet diverse QoS requirements. In contrast, \textit{actor-SAM} and \textit{critic-SAM} show competitive but slightly less consistent results across the slices. The \textit{no-SAM} and \textit{L2-reg} methods lag behind, particularly in managing the latency requirements of slice 3, as depicted by their slower and lower ascents in the CDF. The overall performance trends highlight that while \textit{both-SAM} provides the most comprehensive optimization across the slices, the other approaches such as \textit{actor-SAM} and \textit{critic-SAM}, or the \textit{L2-reg}, offers varying degrees of success in meeting the specific QoS needs, thus affecting the Quality of Experience (QoE) for users across different network interactions.

\subsection{Network Users' QoE }
\begin{figure}[t!]
    \centerline{\includegraphics[width=3.5in]{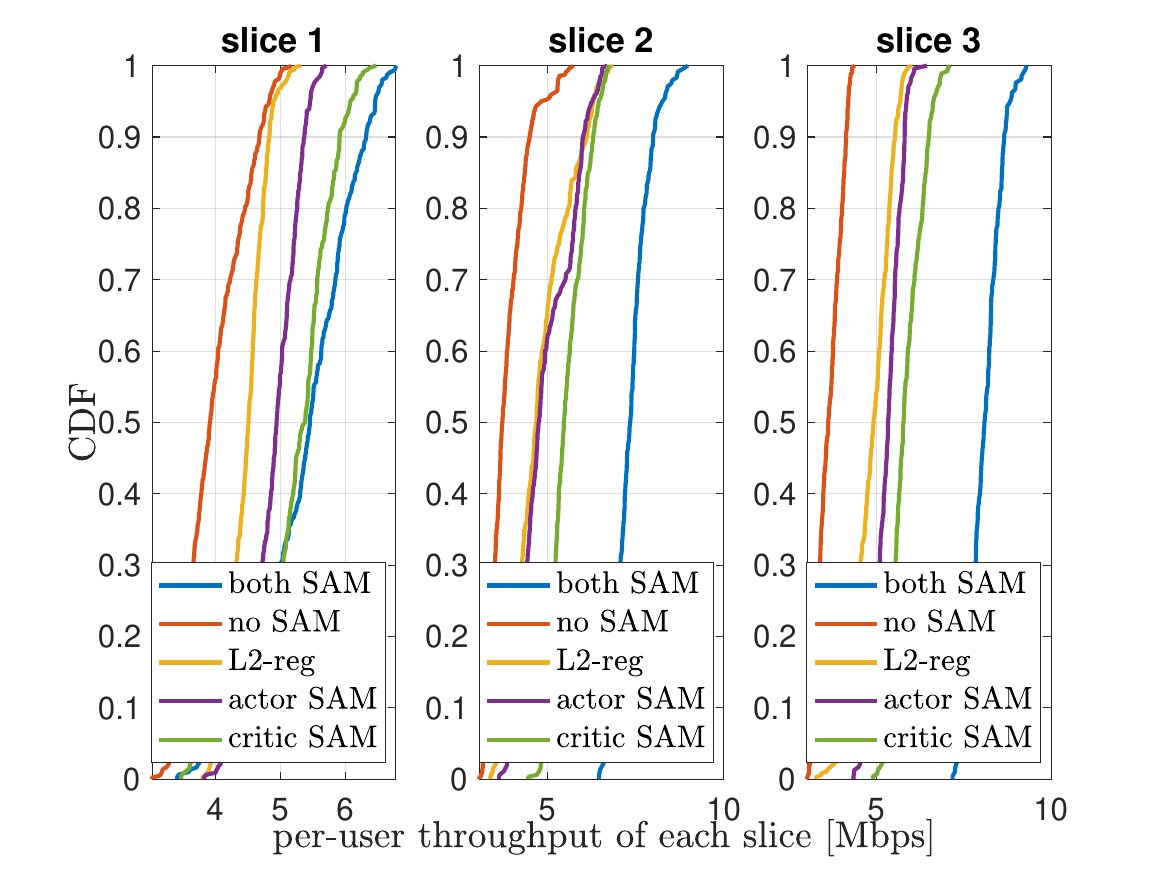}}
    \caption{ CDF of achieved throughput per users across various methods during the training process in a dynamic $\rho$ scenario.}\vspace{-0cm}
    \label{ue-throughput}
    \vspace{-0.cm}
\end{figure}

To show how the improvements in network slices QoS directly enhance user satisfaction, here we show the results and effect of different approaches on UE's throughputs as their QoE. 
Fig. \ref{ue-throughput} represents the CDF of UE's throughput across three network slices, comparing the effectiveness of different methodologies. The graph shows the superior performance of the \textit{both-SAM} method, achieving the highest throughput levels rapidly across all slices, indicative of its robustness in optimizing resource allocation under diverse network conditions. In comparison, \textit{critic-SAM} follows closely, demonstrating consistent performance and ranking second in efficiency, which suggests that applying SAM to the critic might be particularly effective in these scenarios. On the other hand, the \textit{actor-SAM} shows variable outcomes, securing third place with performance that fluctuates more across the slices. The \textit{no-SAM} and \textit{L2-reg} approach lag behind, demonstrating slower progress in the CDF curves and achieving lower throughput peaks, which highlights their relative inefficiency in managing network demands effectively.\vspace{-0.cm}


\begin{table*}[ht]
\centering
\caption{Ablation study on TA-SAM components. Each variant disables one module from the full model.} 
\label{tab:ablation}
\resizebox{\textwidth}{!}{
\begin{tabular}{|l|c|c|c|c|c|c|}
\hline
\textbf{Variant} & \textbf{SAM} & \textbf{Selective (TD-Var)} & \textbf{$\rho$-Schedule} & \textbf{QoS (eMBB)} & \textbf{QoS (mMTC)} & \textbf{QoS (URLLC)} \\
\hline
Full Model (TA-SAM)                    & \checkmark & \checkmark & Adaptive & \textbf{23.67\%} & \textbf{11.19\%}  & \textbf{42.1\%} \\
w/o Selective SAM                      & \checkmark & \(\times\)          & Adaptive &   21.32\%  &  9.81\%   &  38.6\%\\
w/ Static $\rho$                     & \checkmark & \checkmark & Static   &   18.4 \%      &      7.7\%    & 32.06\% \\
w/o SAM (Standard SAC MARL)           & \(\times\)           & \(\times\)           & \(\times\)         &   13.9\%        &   5.87\%         & 25.1\% \\
\hline
\end{tabular}
}
\end{table*}


\begin{table}[ht]
\centering
\caption{Comparison with existing baseline and regularization strategies.}
\label{tab:baslinecomp}
\resizebox{\columnwidth}{!}{
\begin{tabular}{|l|c|c|c|}
\hline
\textbf{Variant} & \textbf{QoS (eMBB)} & \textbf{QoS (mMTC)} & \textbf{QoS (URLLC)} \\
\hline
PLASTIC (adapted)~\cite{lee2024plastic}      &  15.1 \% &   5.9\% & 18.5 \%  \\
SAC + L2 Regularization       & 1.81  \%   &  0.73 \% & 4.8 \% \\
SAC + Dropout (p=0.1)         &  0.67 \%   &  0.4 \% &  2.1\%  \\
SAC + Spectral Normalization  &  3.1 \%   & 1.13 \% &  7.3\%  \\
\hline
\end{tabular}
}
\end{table}

\subsection{Ablation Study and Baseline Comparisons}

To further evaluate the effectiveness of our SAM-based MARL method and isolate the contribution of each component, we conduct an ablation study and benchmark our design against recent regularization strategies as well as the PLASTIC baseline~\cite{lee2024plastic}. Specifically, we examine the following configurations: (1) our proposed method with and without selective SAM regularization, (2) dynamic versus fixed scheduling of the $\rho$ parameter, and (3) the standard SAC MARL baseline. Furthermore, we compare against SAC augmented with common regularization techniques including L2 regularization, Dropout, and Spectral Normalization. These baselines cover the most widely used regularization strategies in deep RL, providing a comprehensive comparison against alternatives beyond L2. We also adapt the PLASTIC framework to our O-RAN setting by incorporating its KL-divergence based policy distillation mechanism. Tables~\ref{tab:ablation} and~\ref{tab:baslinecomp} summarize the results across all configurations.  

As shown in Table~\ref{tab:ablation}, the full TA-SAM model consistently achieves the highest QoS satisfaction across all slices, $23.67\%$ (eMBB), $11.19\%$ (mMTC), and $42.1\% $(URLLC), demonstrating the effectiveness of integrating sharpness aware optimization with task-specific modulation. These results also illustrate the achievable gain over a single centralized DRL agent. Disabling selective SAM regularization results in a drop of $2.35\%–3.5\%$ across the slices, confirming that ignoring agent-specific variance during training hinders adaptive coordination. Replacing the adaptive $\rho$ schedule with a static value leads to a more pronounced degradation, $5–10\%$, emphasizing the importance of dynamically scaling regularization based on task uncertainty. Omitting SAM entirely causes the most significant decline, with the standard SAC MARL baseline yielding only $13.9\%, 5.87\%$, and $25.1\%$ QoS for the three slices, respectively, highlighting the necessity of sharpness aware updates in stabilizing training under dynamic environments.

Further comparison in Table~\ref{tab:baslinecomp} reveals that while traditional regularization methods help mitigate overfitting, they fall short in adapting to O-RAN's dynamic slicing demands. L2 regularization improves SAC performance marginally ($+1.81\%$ for eMBB), while Spectral Normalization performs moderately better ($+3.1\% $for eMBB). Dropout offers the least improvement and in some cases leads to unstable training. In contrast, TA-SAM outperforms the adapted PLASTIC~\cite{lee2024plastic} baseline by $8.5\%$ (eMBB), $5.3\%$ (mMTC), and $23.6\%$ (URLLC), showcasing its superior adaptability and policy refinement in multi slice O-RAN settings. These results validate the necessity of both SAM based optimization and task-aware regularization in achieving robust and generalizable MARL policies.

\section{Conclusions}\label{sec:conclusion}
Managing resources in the dynamic and complex environments of O-RAN architectures presents significant challenges, particularly in terms of maintaining stability and adapting to real-time network conditions. To address these issues, we introduced a novel approach that integrates the SAC algorithm with SAM with applying SAM dynamically based on the variance in TD-errors values. This strategy enhances stability and robustness, particularly for complex tasks. Our approach significantly outperforms non-SAM models, with empirical results showing up to $22\%$ over traditional SAC approach without SAM and a $7\%$ improvement over SAC using L2-regularization, highlighting the effectiveness of SAM in enhancing the generalizability and stability of DRL approaches. This novel resource management model not only addresses the inherent challenges of dynamic resource allocation but also underscores the potential of distributed multi-agent systems in achieving optimal network performance. Future work includes inference oriented deployment in near-RT RICs, where latency, hardware footprint, and accelerator aware policy compression will be key considerations.

\section*{APPENDIX}

\subsection{Notation Table}
To assist the reader, Table~\ref{tab:notation} summarizes key notations and symbols used throughout the manuscript, clearly defining their meanings and contexts.

\begin{table}[ht]
\centering
\caption{Notation Table}
\label{tab:notation}
\begin{tabular}{|c|l|}
\hline
\textbf{Symbol} & \textbf{Description }\\
\hline
$Q^l$ & Aggregate QoS score for slice $l$ \\
$\lambda_i$ & Throughput threshold for UE $i$ \\
$e, b$ & Resource block indices for UEs and slices \\
$\rho$ & SAM perturbation magnitude \\
$\theta$ & Model parameters \\
$\eta$ & Learning rate for parameter updates \\
$a_{b,e}$ & Binary action decision for RB allocation \\
$\gamma$ & Discount factor in reinforcement learning \\
$\sigma(\cdot)$ & Sigmoid activation function \\
$L(\theta)$ & Loss function parameterized by $\theta$ \\
$\nabla_{\theta}$ & Gradient operator w.r.t parameters $\theta$ \\
TD-error & Temporal Difference error \\
$dApp$ & Distributed Application (agent at DU level)\\
$xApp$ & Extended Application (global critic at CU level)\\
CU & Centralized Unit in O-RAN architecture \\
DU & Distributed Unit in O-RAN architecture \\
RIC & RAN Intelligent Controller \\
SINR & Signal-to-Interference-plus-Noise Ratio \\
SAM & Sharpness-Aware Minimization \\
SAC & Soft Actor-Critic algorithm \\
\hline
\end{tabular}
\end{table}

\subsection{Simulation Parameters vs. Standards}
Table~\ref{tab:sim_vs_standards_final} summarizes the key parameters used in our simulation environment and compares them explicitly to corresponding parameters defined by O-RAN Alliance and 3GPP specifications, emphasizing our effort to maintain alignment with standard-compliant scenarios.
\begin{table*}[ht]
\caption{Simulation Settings and Alignment with O-RAN/3GPP Standards}
\label{tab:sim_vs_standards_final}
\centering
\resizebox{1.9\columnwidth}{!}{
\begin{tabular}{|c|c|c|}
\hline
\textbf{Category} & \textbf{Standard/Spec Reference} & \textbf{Our Simulation Setting} \\
\hline
\multicolumn{3}{|c|}{\textbf{Architectural and Interface Compliance}} \\
\hline
System Architecture & O-RAN WG1/WG3 disaggregated RAN & RIC, xApp, and distributed DUs modeled explicitly \\
Control Interfaces & E2 (RIC-to-DU), F1 (DU-to-CU) & Logical abstraction of E2/F1 with modular control delegation \\
E2 Protocol Design & ASN.1-based E2AP and E2SM-RC/Policy & Abstracted message-passing between xApp and DUs \\
E2 Latency Handling & 1–10 ms (typical in O-RAN) & Not explicitly modeled (focus on decision flow, not PHY delay) \\
Slicing Policy Control & A1/E2 Policy Framework & Per-UE slice assignment via central xApp policies \\
Simulator Justification & No public sim supports UE-level RB at DU scale & Custom-built simulator aligned with~\cite{zhang2023neuro} and O-RAN architecture \\
\hline
\multicolumn{3}{|c|}{\textbf{Radio Parameters and RB Granularity}} \\
\hline
Carrier Frequency & 3.5 GHz & 3.5 GHz (urban mid-band) \\
Bandwidth & 100 MHz (FR1, 3GPP TS 38.104) & 100 MHz \\
Subcarrier Spacing & 30 kHz (3GPP TS 38.211) & 30 kHz \\
RB Bandwidth & 180–200 kHz & 200 kHz \\
RB Granularity & Per-UE slice-aware scheduling (O-RAN WG3) & UE-level RB allocation embedded in xApp decision process \\
DU Deployment & Distributed per service area (O-RAN WG1) & 3–7 geographically distributed DUs \\
\hline
\multicolumn{3}{|c|}{\textbf{QoS, Traffic, and Reward Modeling}} \\
\hline
UE Modeling & Real/simulated UEs, USRP-based testbeds & 50–60 synthetic UEs with SINR + dynamic traffic profiles \\
Traffic Generator & TS 23.501-based QoS flows (delay, throughput) & SLA-driven flows with delay/throughput objectives \\
QoS Enforcement & TS 28.530 KPI-based policies & Penalized in reward function for violating delay/QoS targets \\
QoS Metrics & Delay, violation probability, throughput & QoS violation rate, average reward, per-slice metrics \\
\hline
\end{tabular}
}
\end{table*}

\bibliography{Main}
\bibliographystyle{IEEEtran}

\end{document}